\documentclass{article} 
\usepackage[accepted]{icml2021}

\usepackage{amsmath,amsfonts,bm}

\def\eqref#1{equation~\ref{#1}}

\def\1{\bm{1}}

\DeclareMathAlphabet{\mathsfit}{\encodingdefault}{\sfdefault}{m}{sl}
\SetMathAlphabet{\mathsfit}{bold}{\encodingdefault}{\sfdefault}{bx}{n}

\usepackage{hyperref}
\usepackage{url}
\usepackage{microtype}

\usepackage[utf8]{inputenc} 
\usepackage[T1]{fontenc}    
\usepackage{hyperref}       
\usepackage{url}            
\usepackage{booktabs}       
\usepackage{amsfonts}       
\usepackage{nicefrac}       
\usepackage{microtype}      

\usepackage{graphicx}
\usepackage{subfigure}
\usepackage{mathtools}
\usepackage{amsmath}
\usepackage{amssymb}
\usepackage{algorithm}
\usepackage{algorithmic}
\usepackage{dsfont}

\usepackage{xcolor}
\hypersetup{
    colorlinks,
    linkcolor={black},
    citecolor={black},
    urlcolor={black}
}

\icmltitlerunning{Adversarial representation learning for synthetic replacement of private attributes}
\begin{document}

\twocolumn[

\icmltitle{Adversarial representation learning for \\ synthetic replacement of private attributes}

\begin{icmlauthorlist}
\icmlauthor{John Martinsson}{rise}
\icmlauthor{Edvin Listo Zec}{rise}
\icmlauthor{Daniel Gillblad}{aise}
\icmlauthor{Olof Mogren}{rise}
\end{icmlauthorlist}

\icmlaffiliation{rise}{RISE Research Institites of Sweden}
\icmlaffiliation{aise}{AI Sweden}

\icmlcorrespondingauthor{John Martinsson}{john.martinsson@ri.se}

\icmlkeywords{Machine Learning, ICML, Deep Learning, Privacy, Generative Adversarial Privacy}

\vskip 0.3in
]

\printAffiliationsAndNotice{\icmlEqualContribution} 

\begin{abstract}
  Data privacy is an increasingly important aspect of many real-world
  analytics tasks.
  Data sources that contain sensitive information
  may have immense potential
  which could be unlocked using 
  the right privacy enhancing transformations,
  but current methods often fail to produce convincing output.
  Furthermore, finding the right balance
  between privacy and utility is often a tricky trade-off.
  In this work, we propose a novel approach for data privatization,
  which involves two steps: in the first step, it removes the sensitive information,
  and in the second step, it replaces this information with an independent random sample.
  Our method builds on adversarial representation learning
  which ensures strong privacy by training the model to
  fool an increasingly strong adversary.
  While previous methods 
  only aim at obfuscating the sensitive information, we find that
  adding new random information in its place strengthens the
  provided privacy and provides better utility at any given
  level of privacy.
  The result is an approach that can provide stronger privatization
  on image data, and yet be preserving both the domain and the utility
  of the inputs, entirely independent of the downstream task.
\end{abstract}

\section{Introduction}

\begin{figure}[t!]
  \centering
  \includegraphics[width=0.34\textwidth]{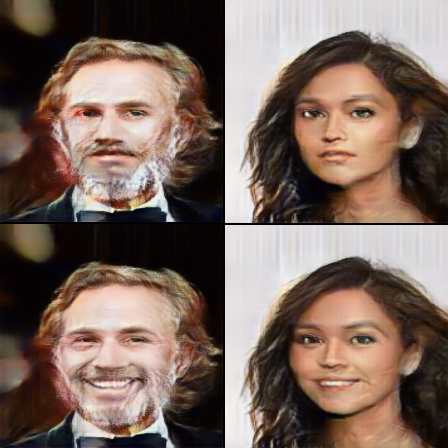}
    \caption{Images censored with the proposed approach.
    Each column corresponds to one input image in Figure~\ref{fig:teaser_images_inputs}.
    The sensitive attribute here is \textit{smiling}.\vspace{-1em}}
  \label{fig:teaser_images_censored}
\end{figure}

Increasing capacity and performance of modern machine learning models
lead to increasing amounts of data required for training them~\citep{goodfellow2016deep}.
However, collecting and using large datasets which may contain sensitive information
about individuals is often impeded by increasingly strong privacy laws
protecting individual rights, and the infeasibility of obtaining individual
consent.
Giving privacy guarantees on a dataset may let us share the data, while protecting the rights of
individuals, and 
thus
unlock
the large benefits for both individuals and the society
that big datasets can provide.

In this work, we propose a technique for selective obfuscation of image datasets.
The aim is to provide the original data as detailed as possible
while making it hard for an adversary
to detect specific sensitive attributes.
The proposed solution is agnostic to the downstream task, 
with the objective to make the data as private as possible
given a distortion constraint.
This issue has previously been addressed
using adversarial representation learning with some success:
a filter model is trained to obfuscate
sensitive information while an adversary model
is trained to recover the information \citep{Edwards_2016}.
In the current work, we demonstrate that \textit{it is easier to hide
sensitive information if you replace it with something else}:
a sample which is independent from the input data.

Aside from the adversary module, our proposed solution includes two main components:
one filter model that is trained to remove the sensitive attribute,
and one generator model that inserts a synthetically generated new
value for the sensitive attribute.
The generated sensitive attribute is entirely independent from
the sensitive attribute in the original input image.
Following previous work in privacy-related adversarial learning
we evaluate the proposed model on faces from the CelebA dataset~\citep{liu2015faceattributes},
and consider, for example, the smile or gender of a person to be the sensitive attribute.
The smile is an attribute that carries interesting aspects 
in the transformations of a human face.
The obvious change reside close to the mouth when a person
smiles, but other subtle changes occur as well:
eyelids tighten, dimples show and the skin wrinkles.
The current work includes a thorough analysis of the dataset,
including correlations of such features.
These correlations make the task interesting and challenging,
reflecting the real difficulty that may occur when anonymizing data.
What is the right trade-off between preserving the utility as defined by allowing
information about other attributes to remain, and removing the 
sensitive information?

In our setup, the adversary can make an arbitrary number of queries to the model. For each query
another sample will be produced from the distribution of the sensitive data,
while keeping as much as possible of the non-sensitive information about
the requested data point.

\begin{figure}[t!]
  \centering
  \includegraphics[width=0.34\textwidth,trim={0 0 0 1.5px},clip]{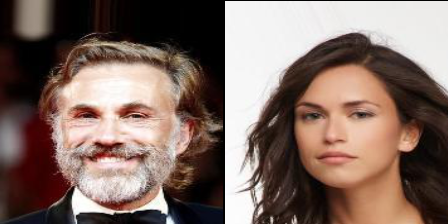}
    \caption{Original uncensored input images censored in Figure~\ref{fig:teaser_images_censored}. The first person has
    bright hair, is male, and is smiling, and the second person has dark hair, is
    female, and is not smiling.\vspace{-1em}}
  \label{fig:teaser_images_inputs}
\end{figure}

Figure~\ref{fig:teaser_images_censored} shows two images censored with the proposed method.
The corresponding input images can be seen in Figure~\ref{fig:teaser_images_inputs}.
The method produce convincing samples from both possible values of the sensitive attribute
(here: \textit{smiling}), while making it hard to determine the true sensitive value.

\section{Related work}
\label{sec:relatedwork}

Privacy-preserving machine learning has been studied from a number of
different angles.
Some work assumes access to a privacy-preserving mechanism,
such as bounding boxes for faces, and studies how to hide people's
identity by blurring~\citep{Oh_2016}, removing~\citep{Orekondy_2018} or
generating the face of other people~\citep{Hukkelas_2019} in their place.
Other work assumes access to the utility-preserving
mechanism and proposes to obfuscate everything except what they want to
retain~\citep{Alharbi_2019}.
This raises the question: how do we find the pixels in an
image that need to be modified to preserve privacy with respect to some
attribute?
\citet{oh2016faceless} 
showed that blurring or removing the head of a person has a limited effect on privacy.
The finding is crucial; \textit{we cannot rely on modifications of an image such
as blurring or overpainting to achieve privacy}. An adversarial set-up instead captures the signals
that the adversary uses, and can attain a stronger privacy.

Adversarial learning is the process of training a model
to fool an adversary
\citep{goodfellow2014generative}.
Both models are trained simultaneously, and become increasingly
good at their respective task during training.
This approach has been successfully used to learn
image-to-image transformations~\citep{Isola_2017,Choi2018}, and synthesis
of properties such as facial expressions~\citep{Song_2017, Tang_2019}.
Privacy-preserving adversarial representation learning
utilize this paradigm to learn
representations of data that hide sensitive information
\citep{Edwards_2016,zhang2018mitigating,xie2017controllable, beutel2017data,raval2017protecting}.

\citet{Bertran_2019} 
minimize the mutual information between the utility variable and the
input image data conditioned on the learned representation.
%
\citet{Roy_2019} 
maximize the entropy of the discriminator
output rather than minimizing the log likelihood, which is beneficial
for stronger privacy.
In the current work, we follow this and include results for both
learning criteria.
%
\citet{Osia_2020} approached the problem using an information-bottleneck. 
%
\citet{Wu_2018}, \citet{Ren_2018}, and \citet{Wang_2019} 
learn transformations of
video that respect a privacy budget while maintaining performance on a
downstream task.
\citet{tran2018representation} proposed an approach for pose-invariant face recognition.
Similar to our work, their approach used adversarial 
learning to disentangle specific attributes in the data.
\citet{oh2017adversarial} trained a model to add a small amount of
noise to the input to hide the identity of a person.
\citet{Xiao2020} 
learn a representation from which it is
hard to reconstruct the original input,
but from which it is possible
to predict a predefined task.
The method provides 
control over which attributes being preserved, but no control over
which attributes being censored. Thus, it puts more emphasis
on preserving utility than privacy, which is not always desired.

All of these, with the exception of \citet{Edwards_2016} (see below), depend
on knowing the downstream task labels. \textit{Our work has no such dependency: the data
produced by our method is designed to be usable regardless of downstream task.} %

%

In \citet{Edwards_2016}, a limited experiment is included which does
not depend on the downstream task. In this experiment, they remove sensitive text 
which was overlaid on images, a task which is much simpler than the real-world
problem considered in the current work. The overlaid text is independent of
the underlying image, and therefore the solution does not require a trade-off
between utility and privacy which is the case in most real settings. Furthermore,
we also replace the sensitive information with synthetic
information which we show further strengthens the privacy.

Like in the current work, \citeauthor{Huang_2017} (\citeyear{Huang_2017}, \citeyear{Huang_2018})
use adversarial learning to minimize the mutual information
between the private attribute and the censored image under a distortion
constraint.
Our solution extends and improves upon these ideas with a modular design consisting of
a filter that is trained to obfuscate
the data, and a generator that further enhances the privacy
by adding new independently sampled synthetic information for the sensitive attributes.
%

\section{Privacy-preserving adversarial representation learning}
\label{sec:method}
In the current work, we propose a novel solution for utility-preserving
privacy-enhancing transformations of data:
we use privacy-preserving representation learning to obfuscate information
in the input data, and output results that retain the information and 
structure of the input.

\subsection{Problem setting}
Generative adversarial privacy (GAP)~\citep{Huang_2018} was proposed as a method
to provide privacy in images under a distortion constraint, and will
be used as the baseline in the current work.
In GAP, one assumes a joint distribution $P(X, S)$ of public data points $X$ and
sensitive private attributes $S$ where $S$ is typically correlated with $X$.
The authors define a privacy mechanism $X' = f(X, Z_1)$ where $Z_1$ is the source of
noise or randomness in $f$. Let $h_f(X')$ be an adversary's
prediction of the sensitive attribute $S$ from the privatized data $X'$
according to a decision rule $h_f$. The performance of the adversary is thus
measured by a loss function $\ell_f(h_f(f(x, z_1)), s)$ and the expected loss
of the adversary with respect to $X$, $S$ and $Z_1$ is
\begin{equation}
  L_{f}(h_f, f) = \mathop{\mathbb{E}}_{\mathclap{\substack{x,s\sim p(x,s) \\
  z_1\sim p(z_1)}}}[\ell_f(h_f(f(x, z_1),
  s)],
\end{equation}
where $p(z_1)$ is the source of noise.

The privacy mechanism $f$ will be trained to be privacy-preserving and
utility-preserving, i.e., it should be hard for an analyst to infer $S$ from
$X'$, but $X'$ should be minimally distorted with respect to $X$. \citet{Huang_2018}
formulate this as a constrained minimax problem
\begin{equation}
  \min_{f}\max_{h_f} -L_{f}(f, h_f)
                   \quad \text{ s.t. } \qquad
                   \mathop{\mathbb{E}}_{\mathclap{\substack{x,s\sim
                   p(x,s)\\z_1\sim p(z_1)}}} [d(f(x, z_1), x)] \leq
                   \epsilon_1,
\label{eq:huang_objective}
\end{equation}
where the constant $\epsilon_1 \geq 0$ defines the allowed distortion for the
privatizer and $d(\cdot, \cdot)$ is some distortion measure.

In the current work, $f$ will be referred to as
the \textit{filter} since the purpose of
it is to filter the sensitive information from $x$. A potential
limitation with this formulation is that it only obfuscates the sensitive information
in $x$ which may make it obvious to the adversary that $x'$ is a censored
version of $x$. Instead, 
we propose to replace the sensitive information with a new independent value $s'$.

\subsection{Our contribution}

We extend the filter with a \textit{generator} module $g$, defined as
$X'' = g(f(X, Z_1), S', Z_2)$ where $S'$ denotes the random variable of the new
synthetic value for the sensitive attribute. $Z_1$ and $Z_2$ denote the sources of
randomness in $f$ and $g$ respectively.
The discriminator $h_g$ is trained to predict $s$ when the input is a real image,
and to predict the ``$fake$'' output when the input comes from $g$ as in
the learning setup in~\citet{Salimans_2016}.
The objective of the generator
$g(x', s', z_2)$ is to generate a new synthetic (independent) sensitive attribute $s'$ in
$x'$, that will fool the discriminator $h_g$. We define the loss of the discriminator $h_g$ as
\begin{equation*}
  \begin{aligned}
  L_{g}(h_g, g) &= \\
      &~ \mathop{\mathbb{E}}_{\mathclap{\substack{x,s\sim p(x,s) \\
        s'\sim p(s') \\
  z_1,z_2\sim p(z_1, z_2) }}} [\ell_g(h_g(g(f(x,z_1), s', z_2)), fake)]
  + \\
  &~ \mathop{\mathbb{E}}_{\mathclap{\substack{x,s\sim p(x,s)}}} [\ell_g(h_g(x), s)],
  \end{aligned}
\end{equation*}
where $p(z_1, z_2)$ is the source of noise, $p(s')$ is the
assumed distribution of the synthetic sensitive attributes $s'$, $fake$ is
the fake class, and $\ell_g$ is the loss function. 
We formulate this as a constrained minimax problem
\begin{equation}
  \begin{aligned}
  \min_{g}\max_{h_g} -&L_{g}(g, h_g) \\
                   \text{ s.t. } \quad
                     &\mathop{\mathbb{E}}_{\mathclap{\substack{x,s\sim
                   p(x,s)\\s'\sim p(s')\\z_1,z_2\sim p(z_1,z_2)}}}[d(g(f(x, z_1), s', z_2),
                   x)] \leq \epsilon_2,
  \end{aligned}
\end{equation}
where the constant $\epsilon_2 \geq 0$ defines the allowed distortion for the
generator. 

\begin{figure*}
  \centering
  \subfigure[]{
  \label{subfig:ce}
  \includegraphics[width=0.26\textwidth]{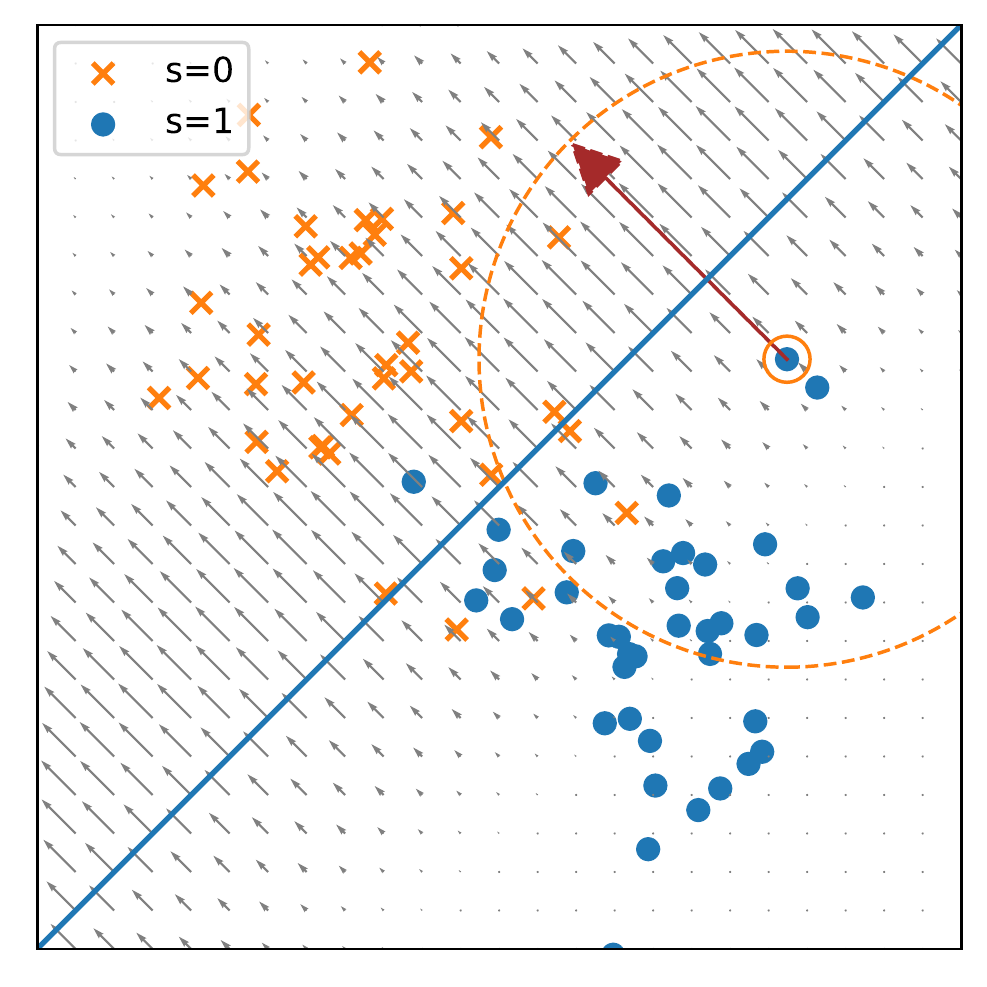}
  }
  \subfigure[]{
  \label{subfig:ent}
  \includegraphics[width=0.26\textwidth]{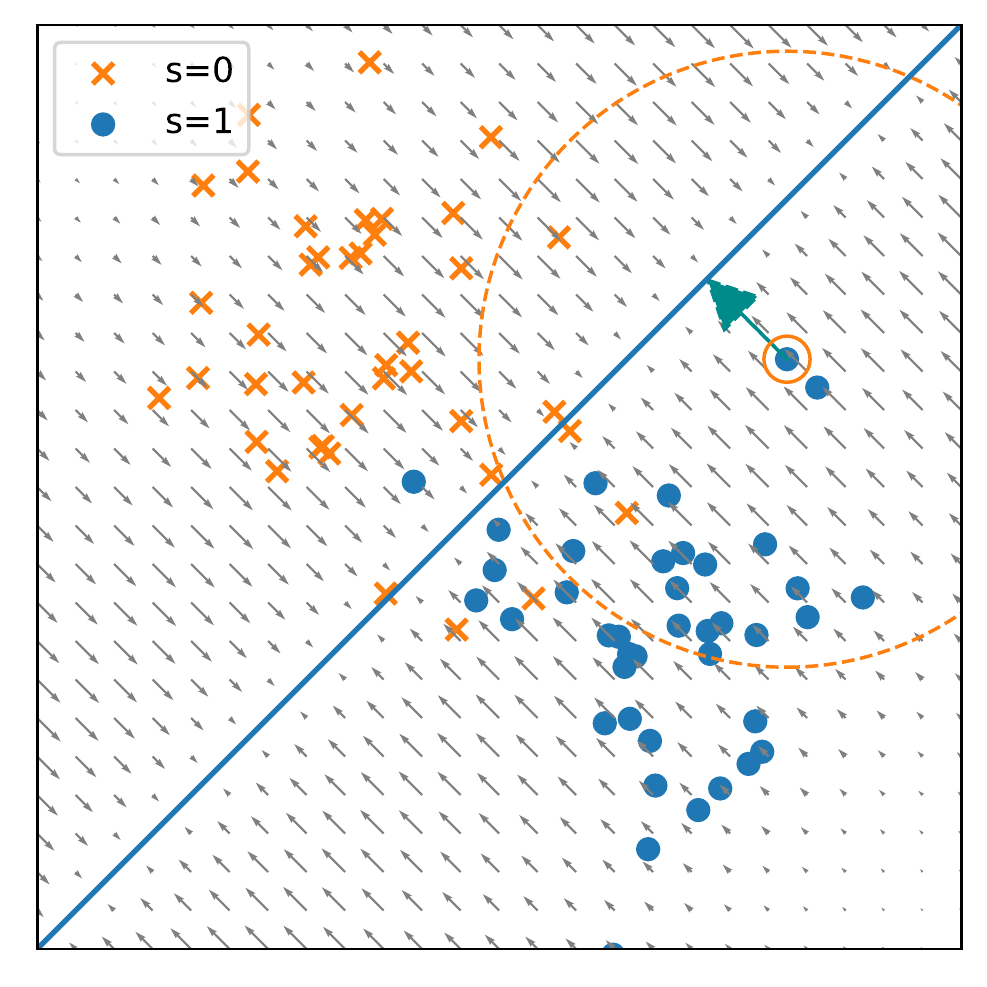}
  }
  \subfigure[]{
  \label{subfig:ours}
  \includegraphics[width=0.26\textwidth]{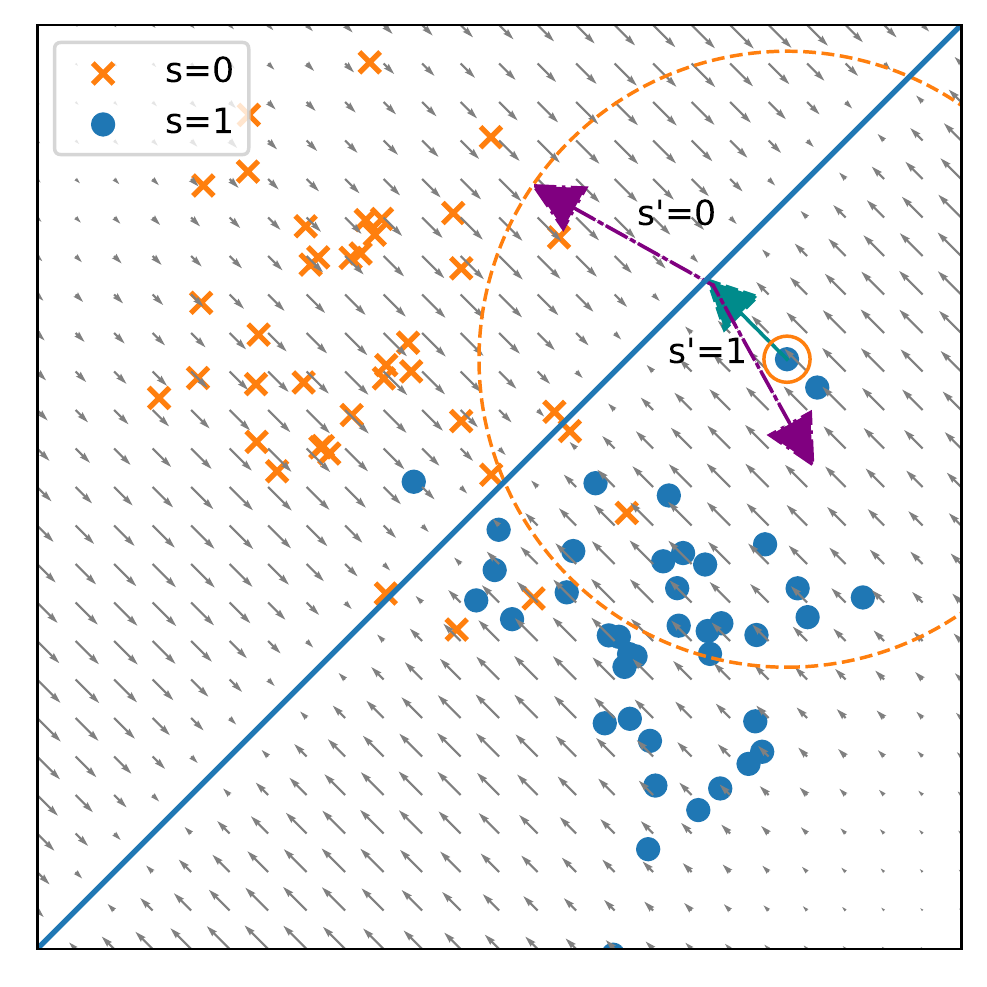}
  }\vspace{-1em}
  \caption{
  Minimizing the log likelihood \subref{subfig:ce} promotes
  transformations of points to the opposite category.
  Maximizing the entropy of the adversary \subref{subfig:ent}
  favors making the adversary uncertain (points closer to the decision boundary).
  Our approach \subref{subfig:ours} 
  first maximizes the entropy as in \subref{subfig:ent},
  and then transforms the image to a random, new value for the
  sensitive attribute $s$ (purple arrows).
  Small grey arrows show the negative gradient of the loss function with regards
  to the input variable, and orange circles illustrates the distortion budget.
  \label{fig:illustration}\vspace{-1em}
  }
\end{figure*}

In Figure~\ref{fig:illustration} we show the difference between \subref{subfig:ce}
minimizing log-likelihood of the adversary, \subref{subfig:ent} maximizing entropy of the
adversary, and \subref{subfig:ours} maximizing the entropy of the adversary and also synthetically
replace the sensitive attribute with a random sample.

\subsection{Implementation}
Let 
$\mathcal{D} = \{(x_i, s_i)\}_{i=1}^{n}$ be a dataset of samples $(x_i, s_i)$ which are assumed
to be identically and independently distributed according to some unknown
joint distribution $P(X, S)$. We assume that the sensitive attribute is binary
and takes values in $\{0, 1\}$. 
However, the proposed approach can easily be extended to categorical sensitive
attributes. 
We model the filter $f(X, Z_1; \theta_f)$ and the generator
$g(X', S', Z_2; \theta_g)$ using convolutional neural networks of the 
UNet architecture~\citep{Ronneberger_2015} parameterized by
$\theta_f$ and $\theta_g$, respectively.
(See Appendix~\ref{additional_details} for details).

The discriminators $h_f(X';\phi_f)$ and
$h_g(X'';\phi_g)$ are modeled using ResNet-18~\citep{He_2016} and a modified
version which we refer to as ResNet-10\footnote{ResNet-10 has the same setup as
  ResNet-18, but each of the ``conv2\_x'', ``conv3\_x'', ``conv4\_x'', and
``conv5\_x'' layers consists of only one block instead of two.},
respectively. The last fully connected layer has been replaced with a two and
three class output layer for each model, respectively.

As suggested by \citet{Roy_2019}, we choose the filter
discriminator loss $\ell_f$ to be the the negative entropy.
Intuitively, this leads to $f$ learning to make the adversary
$h_f$ confused
about the sensitive attribute rather than to make it certain of the wrong value.
For completeness, we also include experiments where
$\ell_f$ is the categorical cross-entropy, as are $\ell_{h_f}$ and $\ell_g$.
The distortion measure $d$ is defined as the
$L2$-norm, and $p(s')$ is assumed to be the uniform distribution $\mathcal{U}\{0
,1\}$. The hyperparameters consist of the learning rate $lr$, the quadratic penalty
term coefficient $\lambda$, the distortion constraint $\epsilon$, and the
$(\beta_1, \beta_2)$ parameters to Adam~\citep{Kingma_2015}. 
%
Details of the training setup can be found in Appendix~\ref{sec:training_setup},
and the full code is published on
GitHub\footnote{https://github.com/anonymous/anonymous-repo-name}. 


\section{Experiments}
\label{sec:experiments}
In this section we describe our experiments, the datasets used,
and the evaluation metrics.

\paragraph{Synthetic data.}
We introduce and apply the method on a synthetic dataset to illustrate the
difference between optimizing the filter for log-likelihoood and entropy, and
why adding the generator is important. The synthetic data consists of
$\mathcal{D}_{train} = \{(x_i, s_i)\}_{i=1}^{n}$, $x\in\mathbb{R}^2$ and
$s\in\{0, 1\}$, drawn from two different normal distributions
$\mathcal{N}(\mu_1, \sigma_1)$ and $\mathcal{N}(\mu_2, \sigma_2)$, where $\mu_1
= [-1, 1]$, and $\sigma_1 = [[0.7, 0] [0, 0.7]]$, and $\mu_2 = [1, -1]$, and
$\sigma_2 = [[0.7, 0] [0, 0.7]]$. Similarily, we construct a test set
$\mathcal{D}_{test} = \{(x_i, s_i)\}_{i=1}^{m}$, and we let $n = 400,000$ and $m
= 2,560$. The points are classified according to which normal distribution they
belong to, and we consider this the secret attribute, i.e, $s=0$ if
$x\sim\mathcal{N}(\mu_1, \sigma_1)$ and, $s=1$ if $x\sim\mathcal{N}(\mu_2,
\sigma_2)$. We sample $s'\sim\mathcal{U}(\{0, 1\})$. We run the experiment with
distortion constraints $\epsilon \in \{0.1, 0.5, 1.0, 1.5, 2.0\}$ and run
each experiment five times.


\paragraph{CelebA.}
The CelebA dataset\footnote{http://mmlab.ie.cuhk.edu.hk/projects/CelebA.html}
\citep{liu2015faceattributes} consists of 202,599 face images of
size 218x178 pixels and 40 binary attribute annotations per image, such as age
(old or young), gender, if the image is blurry, if the person is bald, etc.
The dataset has an official split into a training
set of 162,770 images, a validation set of 19,867 images and a test set of
19,962 images. We resize all images to 64x64 for the quantitative experiments,
and to 128x128 pixels for the qualitative experiments, and normalize all pixel
values to the region $[0, 1]$. We use a higher resolution for the qualitative
results to make subtle visual changes more apparent.

\paragraph{Filtering and replacement of sensitive data.}

Let $\mathcal{D}_{train} = \{(x_i, s_i\}_{i=1}^{n}$ be a set of training data
where $x_i$ denotes facial image $i$ and $s_i\in\{0,1\}$ denotes the sensitive
attribute%
. Further let $\mathcal{D}_{test}
= \{(x_i, s_i)\}_{i=1}^{m}$ be the held out test data.
For the purpose of evaluation, we assume access to a number of utility attributes
$u^{(j)}\in\{0,1\}$ for each $x$ and $j \in U$.
The following attributes provided with the dataset will be used in the CelebA experiments:
\textit{gender}, \textit{lipstick},
\textit{age}, \textit{high cheekbones}, \textit{mouth slightly open},
\textit{heavy makeup}, \textit{smiling}.
In each experiment, one of these will be selected as the sensitive attribute.
The rest will be considered the utility attributes,
not used for training, but allow for a utility score
in the evaluation.
This score shows how well non-sensitive attributes are
preserved in the transformation. We compute an average over
fixed classifiers, trained to predict each respective utility
attribute on the original training data, when evaluated on the censored data.

\paragraph{Hyperparameters.} We train the models using
$\mathcal{D}_{train}$ with $lr=0.0005$, $\lambda=10^5$, $\epsilon\in\{0.03,
  0.02, 0.01, 0.005, 0.001\}$, and $(\beta_1, \beta_2) = (0.9, 0.999)$.
Let the training data censored by the
filter be $\mathcal{D}'_{train} = \{(x'_i, s_i)\}_{i=1}^{n}$, where $x'_i =
f(x_i, z^{(1)}_i;\theta_f)$, and by both the filter and the
generator be $\mathcal{D}''_{train} = \{(x''_i, s_i)\}$ where $x''_i =
g(x'_i,s'_i,z^{(2)}_i;\theta_g)$, $z^{(1)}_i,z^{(2)}_i\sim\mathcal{N}(\mathbf{0}, \mathbf{1})$, and
$s'_i\sim\mathcal{U}\{0,1\}$. We do the same transformations to the test data and
denote them $\mathcal{D}'_{test}$ and $\mathcal{D}''_{test}$ respectively.

\paragraph{Computational requirements.} Each experiment was performed on a Tesla V100 SXM2 32 GB in a DGX-1 machine. The
training was restricted to 100 epochs which takes about 13 hours. We run each
experiment with negative entropy loss three times, and present the average over these
runs, and we run each experiment with log-likelihood loss five times and present
the average over these.

\paragraph{Evaluation.}
To evaluate the privacy loss for our method we train an adversarial classifier
to predict the ground truth sensitive attribute given a training set of
images censored by each privatization method, and then we evaluate each
adversary on a test set of censored images. If an adversary can predict the
ground truth sensitive attribute of censored images, that it has not seen
before, with high accuracy, then the privacy loss is high. Let $adv(s|h(x))$
denote an adversary trained on the censored training set to predict the ground
truth attribute $s$ given censored image $h(x)$ 
then the privacy loss is defined as
\begin{equation}
  \textit{Privacy loss} = \frac{1}{|\mathcal{D}_{test}|}\sum\limits_{(x,s)\in \mathcal{D}_{test}}
  \mathds{1}[adv(s|h(x)) = s].
\end{equation}
To evaluate the utility score on the CelebA data, we use fixed classifiers that have
been pre-trained on the original data to predict a set of utility attributes that are
considered non-sensitive. If these utility attributes are predictable with high
accuracy after privatization of the image, then we consider the method to have a
high utility score. Let $fix(u|x'')$ denote a fixed classifier trained on the
original training set to predict the ground truth attribute $u$, then the
utility score is defined as
\begin{equation}
  \begin{array}{l}
  \textit{(CelebA) }  \text{Utility score} = \\
  \frac{1}{|U|}\sum\limits_{j\in U}\frac{1}{|\mathcal{D}_{test}|}\sum\limits_{(x,u^{(j)})\in \mathcal{D}_{test}}
  \mathds{1}[fix(u^{(j)}|h(x)) = u^{(j)}].
  \end{array}
\end{equation}
To evaluate the utility score on the synthetic data, we compute
\begin{equation}
  \begin{array}{l}
  \textit{(Synthetic) } \text{Utility score} = \\
  1-\frac{1}{\epsilon_{max}|\mathcal{D}_{test}|}\sum\limits_{(x,
  s)\in\mathcal{D}_{test}} |x-h(x)|,
  \end{array}
\end{equation}
one minus the mean distance between original images $x$, and transformed images $h(x)$,
normalized with $\epsilon_{max}=2$ to get distances into the range $[0,
1]$
For the baseline, let $h(x) =
f(x, z_1;\theta_f)$, and for the proposed method, $h(x) = g((f(x, z_1;
\theta_f), s', z_2; \theta_g)$. 

We then plot the utility score against the privacy loss for different distortion
budgets and get a trade-off curve. Similar evaluation method are used
in~\citet{Roy_2019} and \citet{Bertran_2019}. In particular,~\citet{Roy_2019} 
plot accuracy of an adversary predicting the sensitive attribute against the
accuracy of a classifier predicting the target attribute. 

\section{Results}
\label{sec:results}
In this section we present quantitative and qualitative results on the facial
image experiments. 

\begin{figure}[t]
  \centering
  \includegraphics[width=0.5\textwidth]{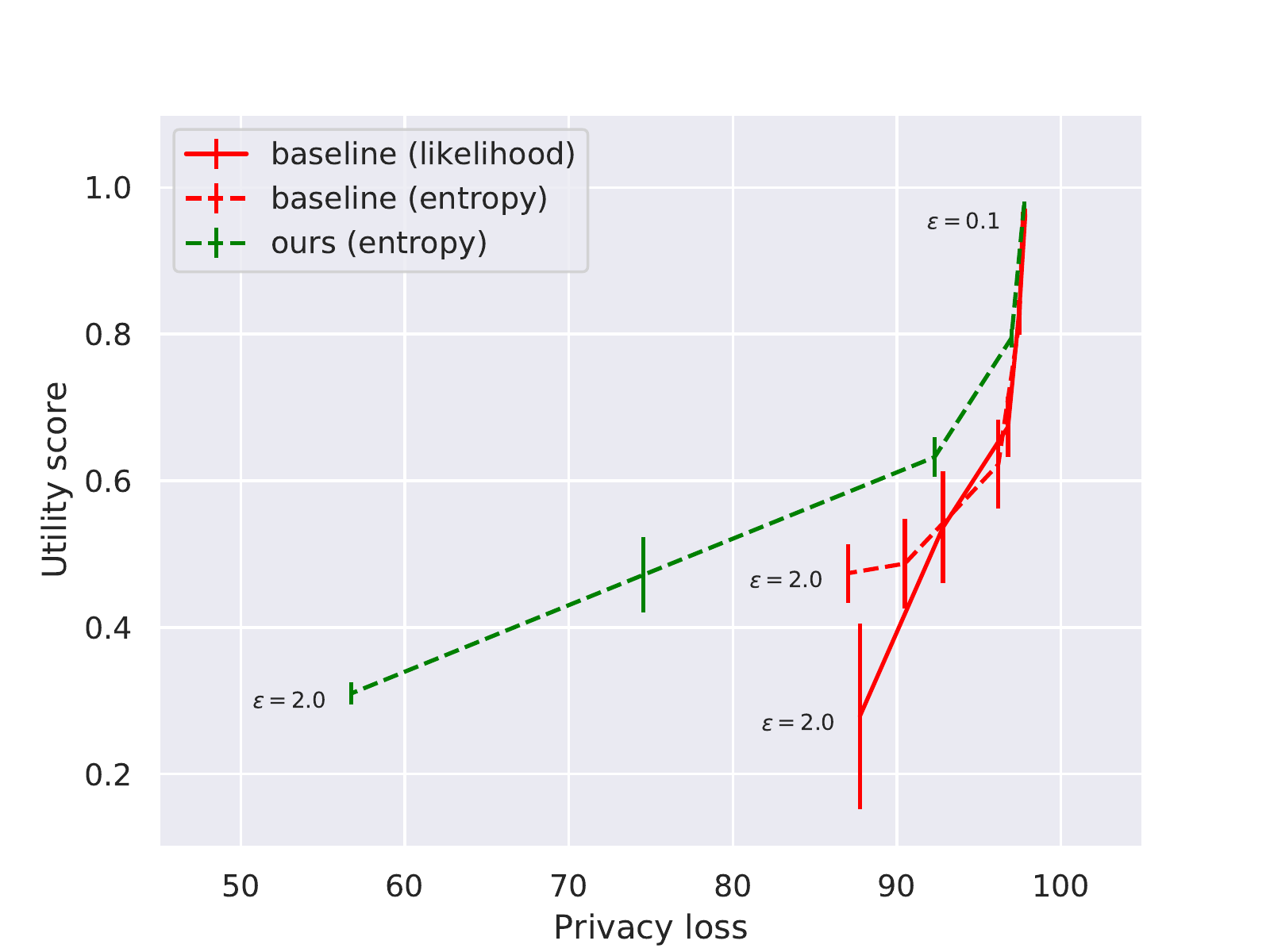}%
  \caption{The trade-off curve between privacy loss and utility score for the
  baseline (likelihood), the improved baseline (entropy) and our method on the
synthetic dataset. Upper left corner is better.}
  \label{fig:privacy_utility_tradeoff_synthetic}
\end{figure}

\begin{figure*}[t]
  \centering
  \subfigure[]{
    \label{subfig:res-loglik}
    \includegraphics[width=.30\textwidth]{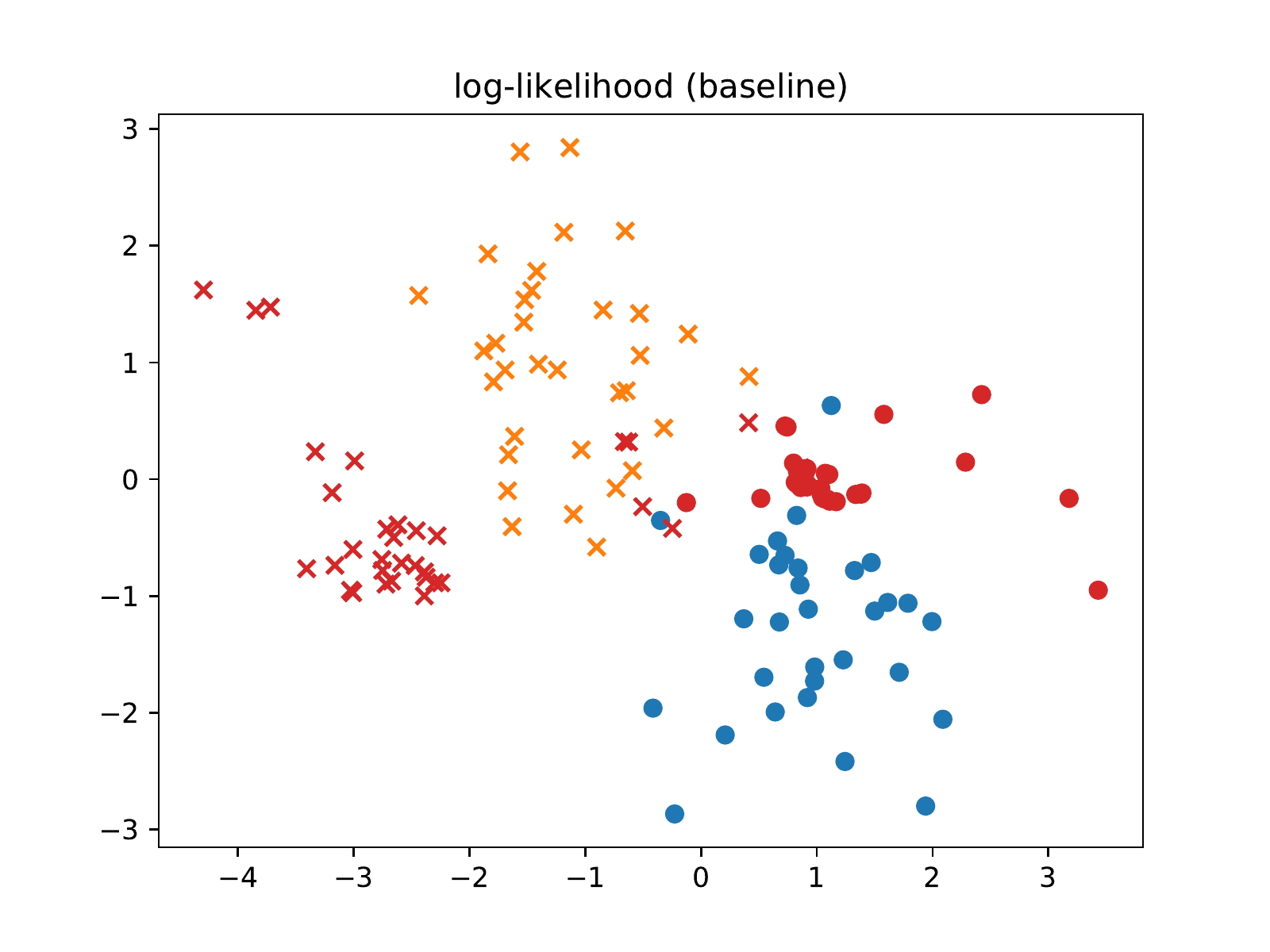}
  }
  \subfigure[]{
    \label{subfig:res-entropy}
    \includegraphics[width=.30\textwidth]{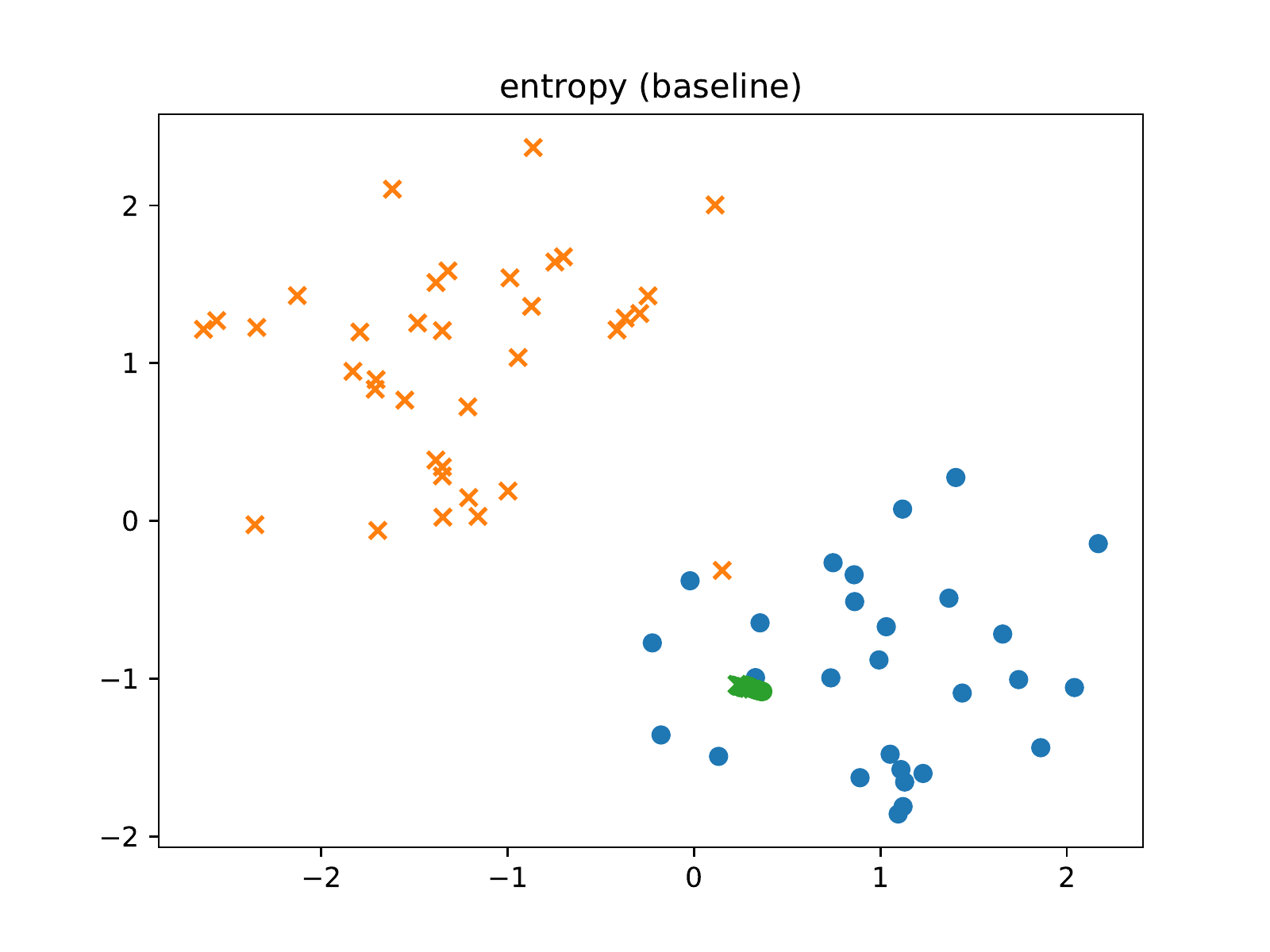}%
  }
  \subfigure[]{
    \label{subfig:res-ours}
    \includegraphics[width=.30\textwidth]{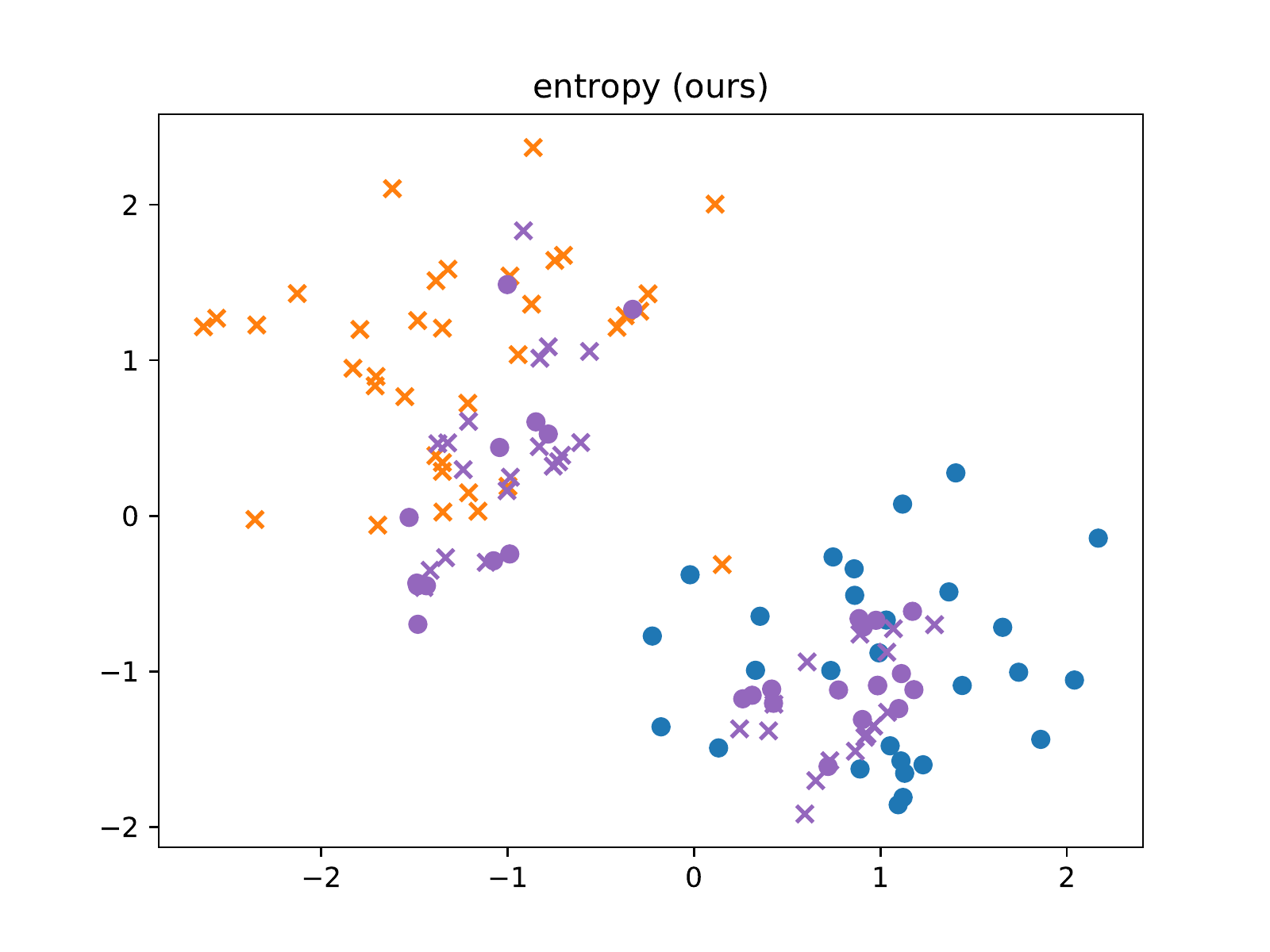}
  }
  \vspace{-1em}
  \caption{Results on the synthetic dataset. The censored representations by
    \subref{subfig:res-loglik} baseline
    with filter trained to minimize log-likelihood of adversary,
    \subref{subfig:res-entropy} the baseline filter trained to
    maximize entropy, and 
    \subref{subfig:res-ours}  
    filter trained to maximize entropy and then generating a
    new synthetic attribute.}
  \label{fig:representations}
\end{figure*}

\paragraph{Synthetic data.}
Figure~\ref{fig:privacy_utility_tradeoff_synthetic} shows the trade-off between
privacy loss and utility score for the experiment on the synthetic dataset. Our method
consistently outperforms the baseline for all given distortion budgets.
%
Figure~\ref{fig:representations} shows the original data (orange for s=0, and
blue for s=1) together with the censored data for the three different transforms
(red for baseline with likelihood, green for baseline with entropy, and purple
for the proposed method). When minimizing log-likelihood we see clear clusters on
opposite side of the decision boundary, when maximizing entropy we see that all
points approach the decision boundary, and when adding the generator we see that
the points are mapped, at random, into the cluster of orange points or blue
points depending on the sampled synthetic sensitive attribute.

\begin{figure*}[t!]
  \centering
  \includegraphics[width=.45\textwidth]{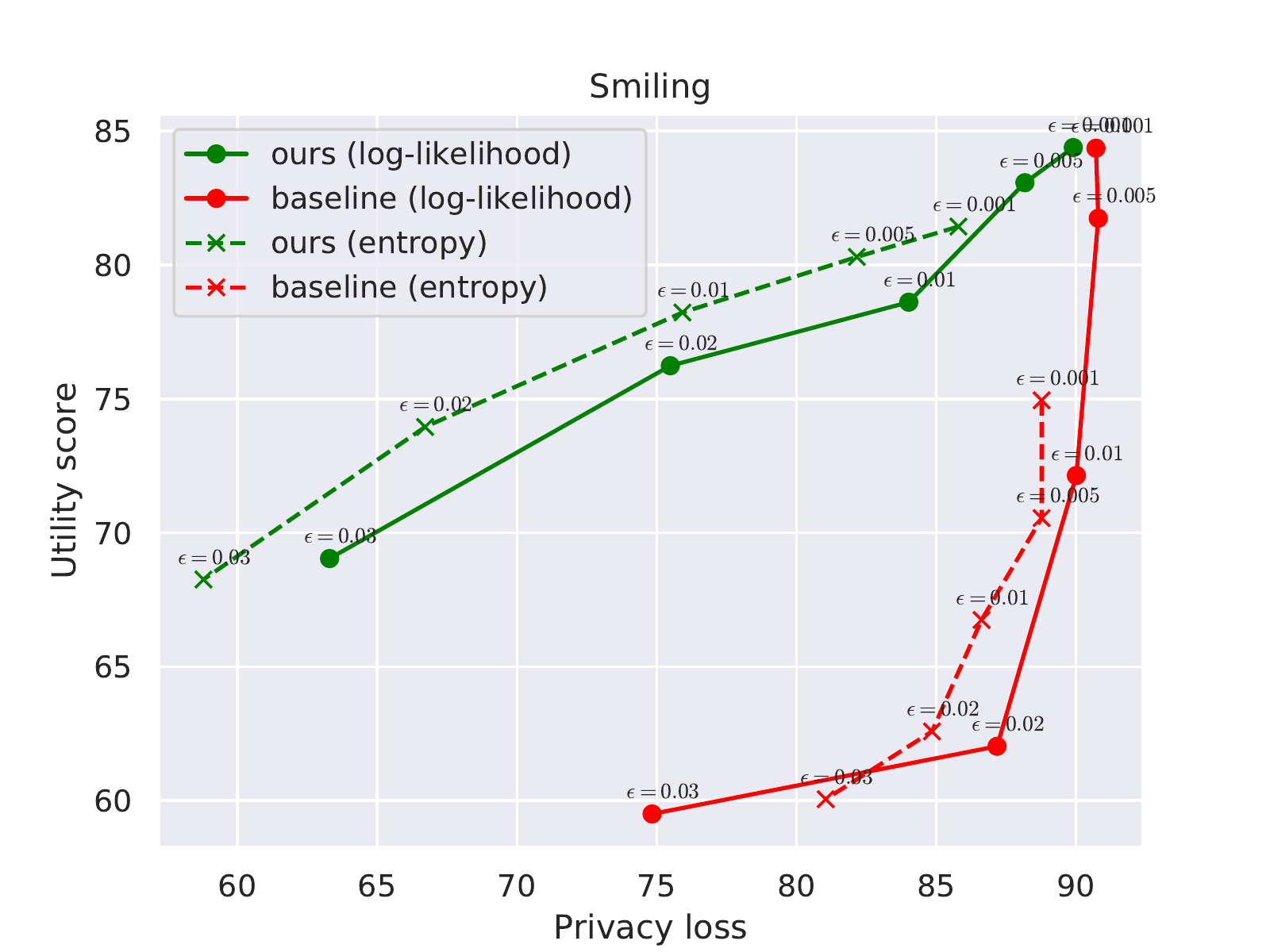}%
  \includegraphics[width=.45\textwidth]{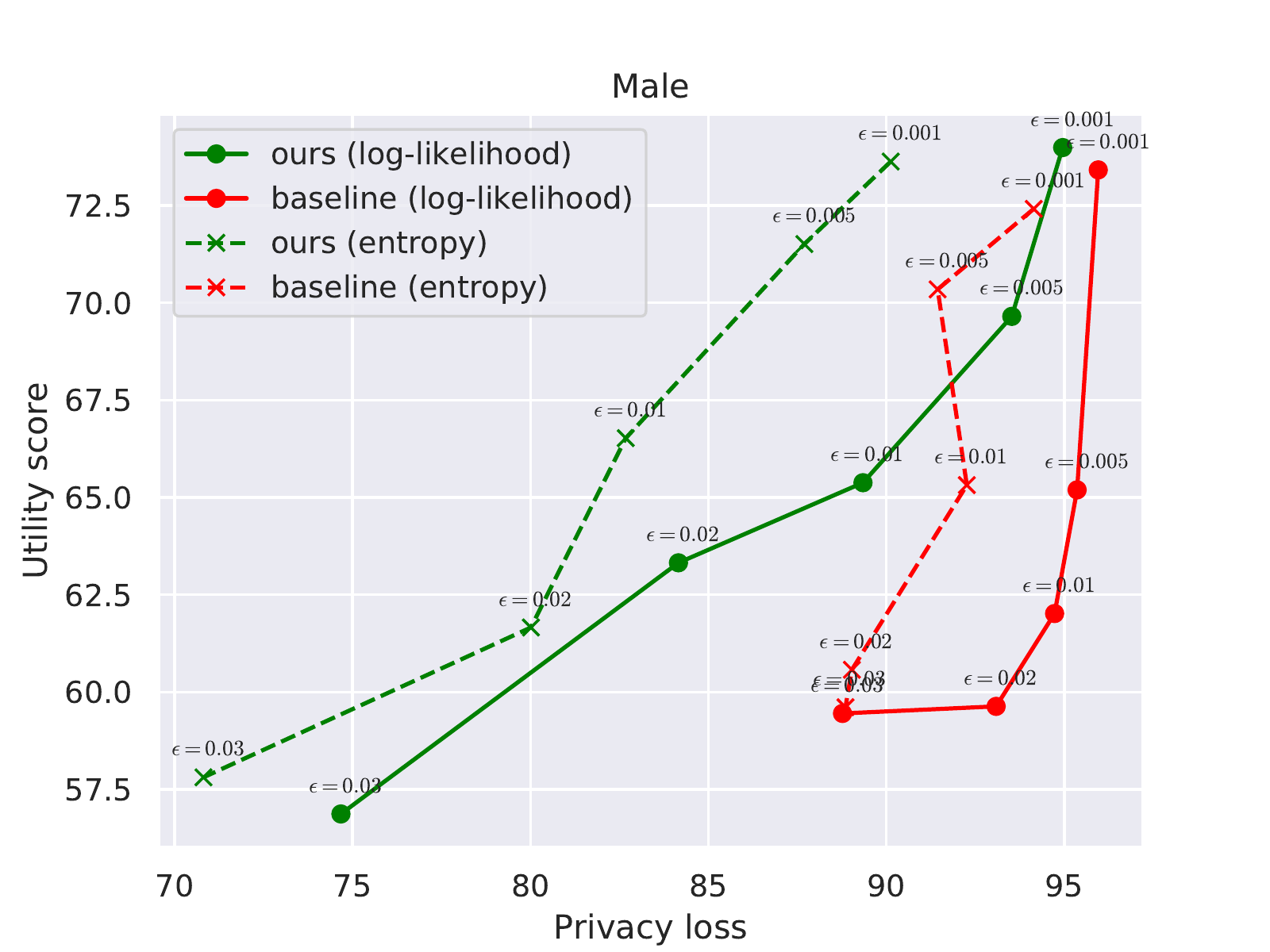}
  \includegraphics[width=.45\textwidth]{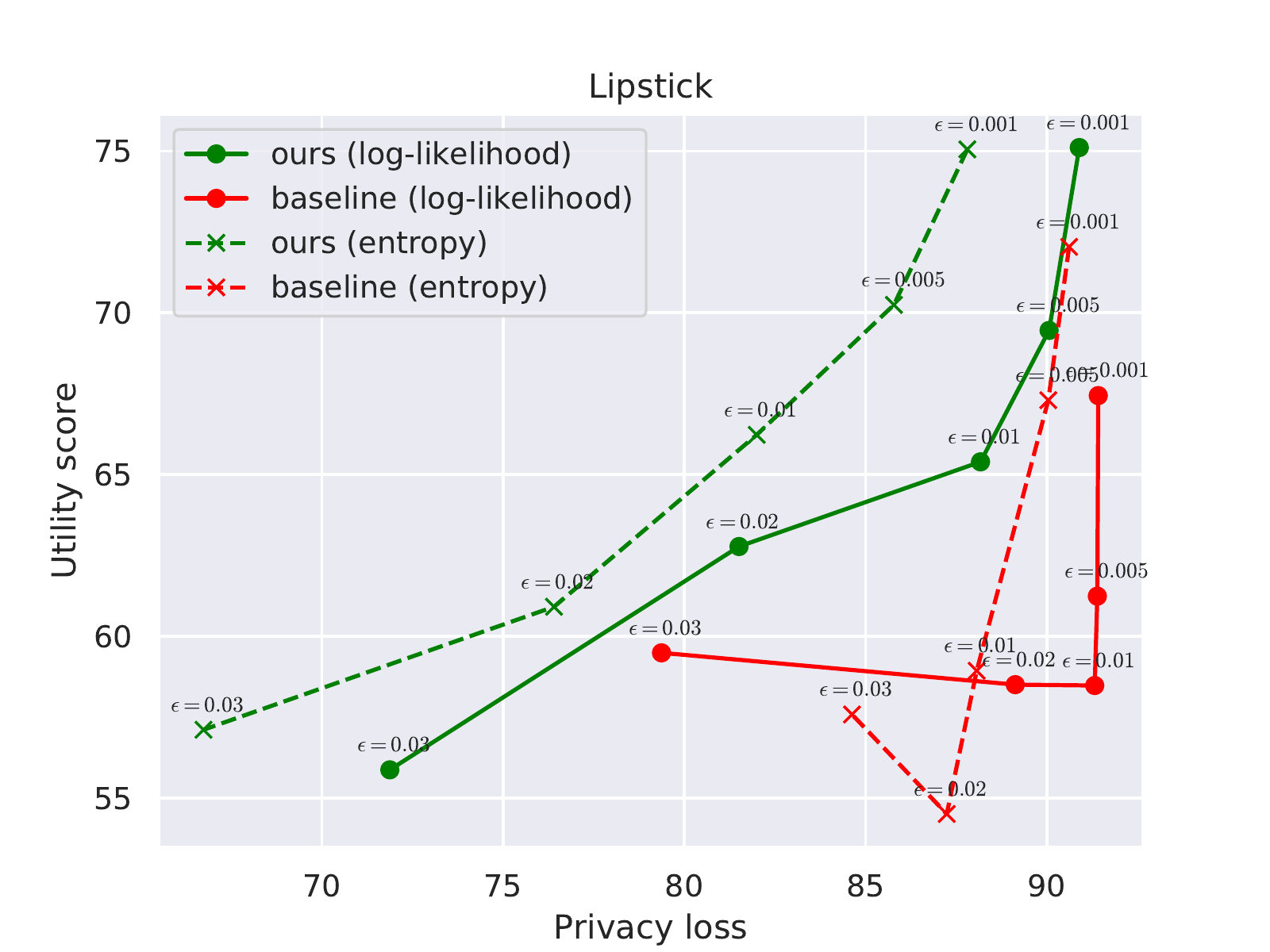}%
  \includegraphics[width=.45\textwidth]{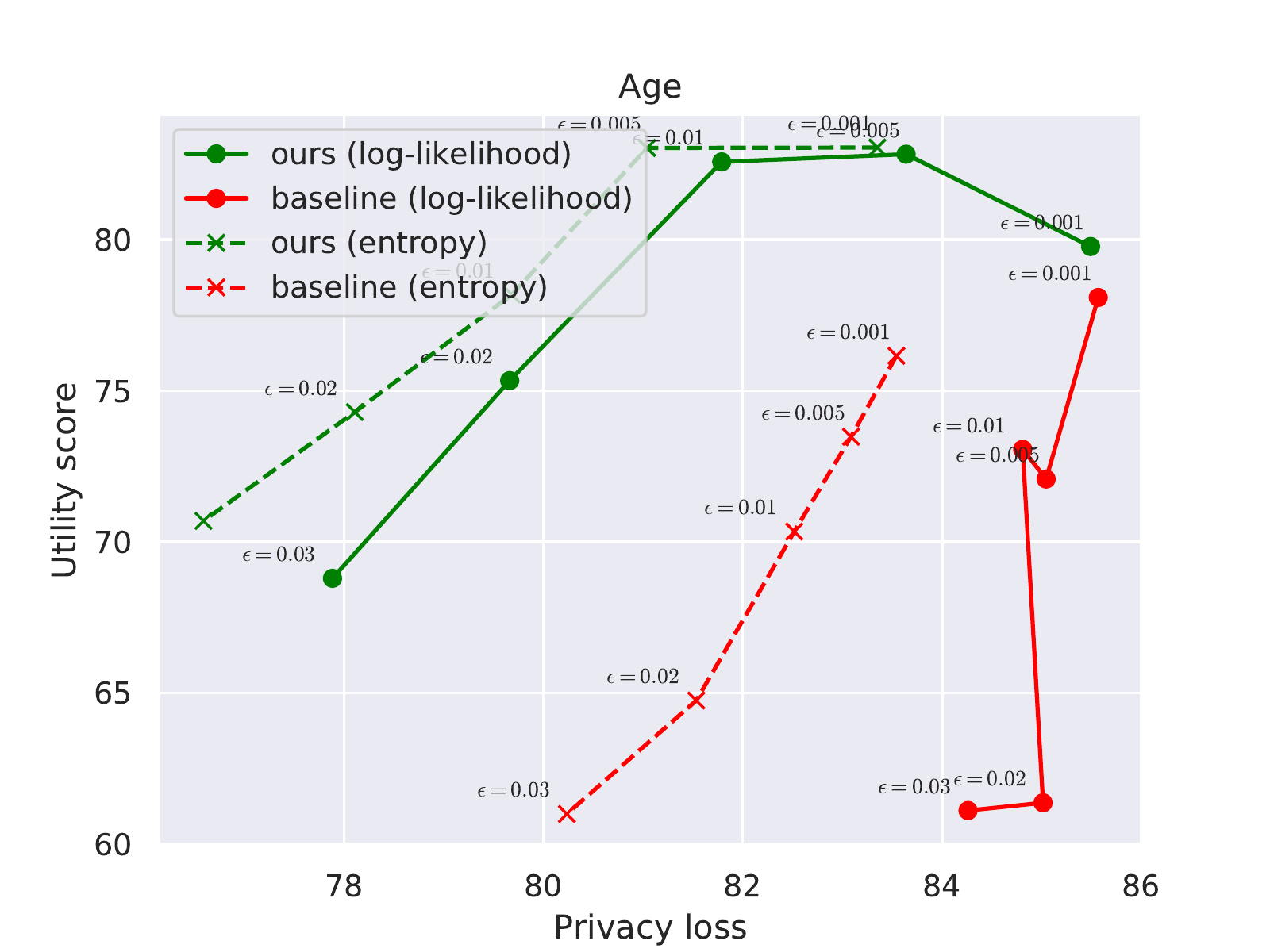}
  \caption{Privacy vs. utility trade-off curve where the sensitive attribute is
    \textit{smiling} (top left), \textit{gender} (top right), \textit{lipstick}
    (bottom left), \textit{age} (bottom right). Our approach with
negative entropy loss consisitently outperforms all other approaches on the
attributes explored, and our method with log-likelihood loss 
outperforms the baseline with log-likelihood loss on all explored attributes.} 


  \label{fig:privacy_utility_tradeoff_1}
\end{figure*}

\paragraph{CelebA.}
%
Figure~\ref{fig:privacy_utility_tradeoff_1} shows the trade-off between privacy
loss and utility score
when evaluated on censored images.
Our method consistently has a higher
utility at any given level of privacy compared to the baseline.
Remember: these are strong adversaries required to run tagged training data through the privacy
mechanism to be able to train. Additional results can be found in 
Appendix~\ref{additional_results}.

To further show that explicitly optimizing for privacy in the privatization mechanism
is necessary we have conducted a similar experiment using
StarGAN~\cite{Choi2018}
to randomly change the sensitive attribute in the image (results in suppl.). We
then evaluate the censored images,
which look very convincing to a human, using adversarial classifiers. 
The adversaries can successfully detect the sensitive attributes with an accuracy of roughly 90\%.
For the weight $\lambda_{rec}$ of the cycle
consistency loss we explored values $\{0,5,10,50\}$. We obtain similar scores when
we exclude the filter part of our method and use only the generator part to censor the images (see
Appendix~\ref{additional_results}).

\begin{table}[]
  \caption{The success rate of our method to fool a
		fixed classifier for the sensitive attributes \textit{smiling}, \textit{gender},
    \textit{lipstick}, and \textit{age}. 
    This was measured as the accuracy of $fix(s|\cdot)$ for each censored attribute in
    the censored data $\{(x''_i, s'_i)\}_{i=1}^{m}$. 
    Higher is better. Average and standard deviation of five runs.
    }
  \label{tab:results_4}
\begin{center}
\setlength{\tabcolsep}{4pt} 
\begin{tabular}{lllll}
  \textbf{Dist.} &\multicolumn{3}{c}{\textbf{Synthetic}}  \\
	$\epsilon$ & Smiling & Gender & Lipstick & Young \\
\hline \\
 0.001 & $82.4 \pm 2.1$ & $72.9 \pm 1.5$ & $62.2 \pm 3.1$ & $59.2 \pm 1.9$ \\
 0.005 & $86.4 \pm 3.7$ & $79.4 \pm 0.6$ & $71.7 \pm 2.4$ & $62.1 \pm 2.5$ \\
  0.01 & $87.4 \pm 2.1$ & $85.6 \pm 1.4$ & $77.6 \pm 2.4$ & $59.6 \pm 1.6$ \\
  0.05 & $91.2 \pm 2.9$ & $90.3 \pm 4.3$ & $90.6 \pm 4.2$ & $67.7 \pm 2.1$ \\
\end{tabular}
\vspace{-2em}
\end{center}
\end{table}
In Table~\ref{tab:results_4} we present the results of evaluating the accuracy
of $fix(s|\cdot)$ on the dataset $\{x''_i,
s'_i\}_{i=1}^{m}$ where $x''_i = g(x'_i, s'_i z^{(2)}_i)$ is the image censored
with our method and $s'_i$ is the new synthetic attribute uniformly sampled from
$\{0, 1\}$. This means that we measure how often the classifier predict the new
synthetic attribute $s'_i$ when applied to $x''_i$. We can see that with
$\epsilon=0.001$ the method is on average able to fool the classifier $82.4\%$
of the time for the \textit{smiling} attribute,
and this increases with larger distortion budget $\epsilon$ to a success rate of $91.2\%$ on
average with
$\epsilon=0.05$. The results are similar when the images have been
censored with respect to the attributes \textit{gender}, \textit{lipstick}, and
\textit{age}, but these require a larger distortion budget.

\begin{figure*}[]
  \centering
  \includegraphics[width=1.0\textwidth]{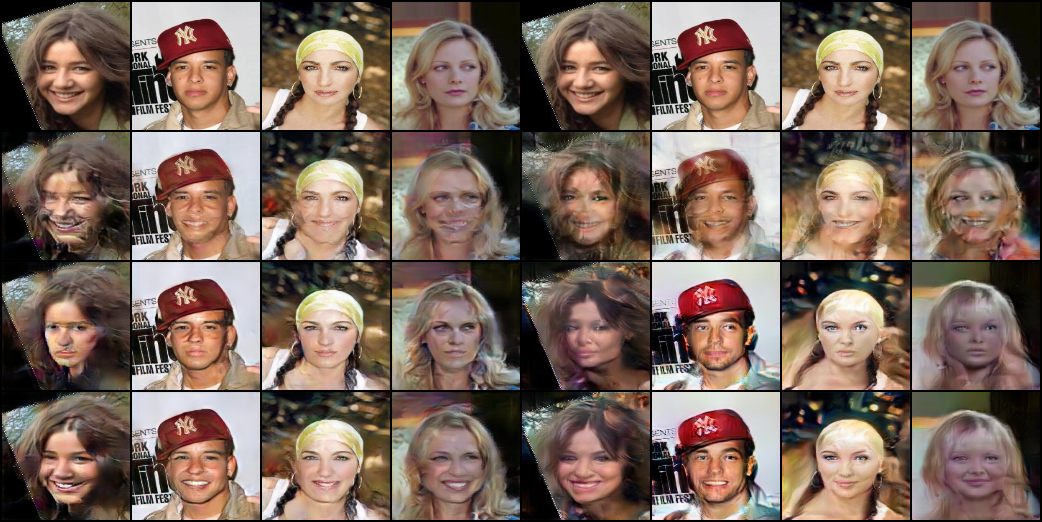}
  \caption{Qualitative results for the sensitive attribute smile. In the first four columns: $\epsilon=0.001$,
    and  in the last four columns: $\epsilon=0.01$. From top to bottom row: input image ($x$), censored image ($x'$), censored image
    with synthetic non-smile ($x'', s'=0$), censored image with synthetic smile 
    ($x'', s'=1$). The model is able to generate a synthetic smiling attribute 
	while maintaining much of the structure in the image. These images were
generated from a model trained using 128x128 pixels.} 
  \label{fig:qualitative_1}
\end{figure*}

Figure~\ref{fig:qualitative_1} shows, from the top row to the bottom row, the
input image $x$, the censored image $x'$, the censored image $x''$ with the synthetic
attribute $s'=0$ (non-smiling), and the censored image $x''$ with the synthetic
attribute $s'=1$ (smiling). A value of $\epsilon=0.001$ is used in the first four columns, and
$\epsilon=0.01$ in the last four columns. The images
censored by our method look sharper and it is less obvious that they are
censored. We can see that the method convincingly generates non-smiling
faces and smiling faces while most of the other parts of the image
is intact. These images are sampled from models trained on images of 128x128 pixels
resolution. See Figure~\ref{fig:qualitative_2} in
Appendix~\ref{additional_results} for corresponding
samples on the same input images, but using \textit{gender}
as the sensitive attribute.



\section{Discussion}
\label{sec:discussion}

The experimental results suggest that we can obtain a stronger privacy
using the proposed two-step process.

In Figure~\ref{fig:illustration}, we show 
how a privatization mechanism is affected by an adversarial setup and its
training criterion. Minimizing the log likelihood of the adversary can be interpreted
as attempting to make the adversary certain of the wrong value of the sensitive attribute
and may lead to transforming data with a certain sensitive value to be transformed to similar outputs.
Using the entropy loss instead
leads to the adversary becoming less certain and thus leads to more privacy.
Using this approach to remove sensitive information, and then adding random
information ensures an even stronger privacy, as we take another step adding
randomness to the output.
This intuition is confirmed in the experimental results on the synthetic data,
as demonstrated by Figure~\ref{fig:representations}. Our approach
successfully transforms the datapoints into a distribution which is nearly
indistinguishable from the input distribution, and where datapoints with the
different sensitive values are completely mixed.

Our method consistently outperforms the baseline, ensuring a higher level of
privacy while retaining more of the utility (see
Figure~\ref{fig:privacy_utility_tradeoff_1}). For all sensitive attributes that
we consider and at nearly all given distortion budgets ($\epsilon \in \{0.02,
0.01, 0.005, 0.001\}$) we observe both a higher privacy and a higher utility
for our method. For some of the attributes (\textit{gender}, \textit{lipstick}),
at extreme distortions
($\epsilon = 0.03$) our method has a utility score which is not better
than the baseline,
but still provide considerably better privacy.
Furthermore, we can observe that using the
entropy loss function for the filter benefits both the baseline and our method.


This shows that our method makes it more difficult for the adversary to see
through the privatization step at each given distortion budget. To show the
effect of the filter we have conducted experiments where we only use the
generator to privatize the images (see Appendix~\ref{additional_results}), which
does not seem to provide any privacy.

A generator without the filter may learn to either pass the original image without
modification (when $s' = s$) or to transform the image into a random other value ($s' \neq s$).
If the transformed image is indistinguishable from a real image this is not a
problem, but otherwise we can easily reverse the privatization by detecting
if the image is real or not. The filter step mitigates this by \textit{always}
removing the sensitive data in the image, forcing the generator to synthesize
new data. Since the censored image is now guaranteed to be synthetic, we are no
longer susceptible the simple real/fake attack.

Similarily, we conduct experiments where we use StarGAN to censor the images, but
observe no privacy when using this standard attribute manipulation
method. A reason could be that the cyclic consistency loss term actually
encourage image transformations that are easily invertible given the original
attribute, which is not desirable in a privacy setting.  This motivates our
approach which is explicit about the privacy objective.


In Table~\ref{tab:results_4}, we see that the fixed smile classifier
is fooled by our privatization mechanism in 82.4\% to 91.2\% of the data points
in the test set (depending on the distortion $\epsilon$).
These results indicate that it may be harder for an adversarially trained classifier to
predict the sensitive attribute when it has been replaced with something
else, as compared to simply removed. We assume that this is due to
the added variability in the data. Or intuitively: it is easier to
``blend in'' with other images that have similar demonstrations of 
the sensitive attribute, such as the smile.


The fact that many important attributes in facial images correlate leads to
the reflection that disentangling the underlying factors of variation is
not entirely possible. For example, in CelebA, lipstick is highly correlated with female gender. 
This means that if we want to hide all information about the presence of
lipstick we also need to hide the gender (and other
correlating attributes). This problem is further analysed in
Table~\ref{tab:results_5} in Appendix~\ref{additional_results}.

A strength of our method is that it is \textit{domain-preserving},
this allows a utility provider
to use the censored image in existing algorithms without modifications.
Since the method also preserves utility this may also allow stacking the
privatization mechanism to censor multiple attributes in an image.

\section{Conclusions}

\label{sec:conclusions}
In this work we have presented a strong privacy-preserving
transformation mechanism for image data which is learned using an adversarial setup.
While previous work on adversarial representation learning has focused on 
removing information from a representation, our
approach extends on this and can also generate new information in its place that looks
realistic and gives further privacy properties compared to the baseline.
This new information is sampled independently from the input and realised using
the adversarially trained generator module.
We evaluate our method using adversarially trained classifiers, and our results show
that not only do we provide stronger privacy with regards to sensitive attributes;
we also preserve non-sensitive attributes of the image at a higher rate.
In this very challenging evaluation setting, the adversary 
is allowed to be trained using tagged data and the output of the privacy mechanism,
highlighting the strength of our results.

\bibliography{references}

\begin{thebibliography}{32}
\providecommand{\natexlab}[1]{#1}
\providecommand{\url}[1]{\texttt{#1}}
\expandafter\ifx\csname urlstyle\endcsname\relax
  \providecommand{\doi}[1]{doi: #1}\else
  \providecommand{\doi}{doi: \begingroup \urlstyle{rm}\Url}\fi

\bibitem[Alharbi et~al.(2019)Alharbi, Tolba, Petito, Hester, and
  Alshurafa]{Alharbi_2019}
Alharbi, R., Tolba, M., Petito, L.~C., Hester, J., and Alshurafa, N.
\newblock To mask or not to mask? balancing privacy with visual confirmation
  utility in activity-oriented wearable cameras.
\newblock \emph{Proc. ACM Interact. Mob. Wearable Ubiquitous Technol.},
  3\penalty0 (3), September 2019.
\newblock \doi{10.1145/3351230}.
\newblock URL \url{https://doi.org/10.1145/3351230}.

\bibitem[Bertran et~al.(2019)Bertran, Martinez, Papadaki, Qiu, Rodrigues,
  Reeves, and Sapiro]{Bertran_2019}
Bertran, M., Martinez, N., Papadaki, A., Qiu, Q., Rodrigues, M., Reeves, G.,
  and Sapiro, G.
\newblock Adversarially learned representations for information obfuscation and
  inference.
\newblock In Chaudhuri, K. and Salakhutdinov, R. (eds.), \emph{Proceedings of
  the 36th International Conference on Machine Learning}, volume~97 of
  \emph{Proceedings of Machine Learning Research}, pp.\  614--623, Long Beach,
  California, USA, 09--15 Jun 2019. PMLR.

\bibitem[Beutel et~al.(2017)Beutel, Chen, Zhao, and Chi]{beutel2017data}
Beutel, A., Chen, J., Zhao, Z., and Chi, E.~H.
\newblock Data decisions and theoretical implications when adversarially
  learning fair representations.
\newblock \emph{arXiv preprint arXiv:1707.00075}, 2017.

\bibitem[Choi et~al.(2018)Choi, Choi, Kim, Ha, Kim, and Choo]{Choi2018}
Choi, Y., Choi, M., Kim, M., Ha, J.~W., Kim, S., and Choo, J.
\newblock {StarGAN: Unified Generative Adversarial Networks for Multi-domain
  Image-to-Image Translation}.
\newblock \emph{Proceedings of the IEEE Computer Society Conference on Computer
  Vision and Pattern Recognition}, pp.\  8789--8797, 2018.
\newblock ISSN 10636919.
\newblock \doi{10.1109/CVPR.2018.00916}.

\bibitem[Edwards \& Storkey(2016)Edwards and Storkey]{Edwards_2016}
Edwards, H. and Storkey, A.~J.
\newblock Censoring representations with an adversary.
\newblock In \emph{4th International Conference on Learning Representations,
  {ICLR} 2016, San Juan, Puerto Rico, May 2-4, 2016, Conference Track
  Proceedings}, 2016.

\bibitem[Goodfellow et~al.(2014)Goodfellow, Pouget-Abadie, Mirza, Xu,
  Warde-Farley, Ozair, Courville, and Bengio]{goodfellow2014generative}
Goodfellow, I., Pouget-Abadie, J., Mirza, M., Xu, B., Warde-Farley, D., Ozair,
  S., Courville, A., and Bengio, Y.
\newblock Generative adversarial nets.
\newblock In \emph{Advances in neural information processing systems}, pp.\
  2672--2680, 2014.

\bibitem[Goodfellow et~al.(2016)Goodfellow, Bengio, and
  Courville]{goodfellow2016deep}
Goodfellow, I., Bengio, Y., and Courville, A.
\newblock \emph{Deep learning}.
\newblock MIT press, 2016.

\bibitem[{He} et~al.(2016){He}, {Zhang}, {Ren}, and {Sun}]{He_2016}
{He}, K., {Zhang}, X., {Ren}, S., and {Sun}, J.
\newblock Deep residual learning for image recognition.
\newblock In \emph{2016 IEEE Conference on Computer Vision and Pattern
  Recognition (CVPR)}, pp.\  770--778, June 2016.
\newblock \doi{10.1109/CVPR.2016.90}.

\bibitem[Huang et~al.(2017)Huang, Kairouz, Chen, Sankar, and
  Rajagopal]{Huang_2017}
Huang, C., Kairouz, P., Chen, X., Sankar, L., and Rajagopal, R.
\newblock Context-aware generative adversarial privacy.
\newblock \emph{Entropy}, 19\penalty0 (12), 2017.
\newblock ISSN 1099-4300.
\newblock \doi{10.3390/e19120656}.
\newblock URL \url{https://www.mdpi.com/1099-4300/19/12/656}.

\bibitem[{Huang} et~al.(2018){Huang}, {Kairouz}, and {Sankar}]{Huang_2018}
{Huang}, C., {Kairouz}, P., and {Sankar}, L.
\newblock Generative adversarial privacy: A data-driven approach to
  information-theoretic privacy.
\newblock In \emph{2018 52nd Asilomar Conference on Signals, Systems, and
  Computers}, pp.\  2162--2166, Oct 2018.
\newblock \doi{10.1109/ACSSC.2018.8645532}.

\bibitem[Hukkel{\aa}s et~al.(2019)Hukkel{\aa}s, Mester, and
  Lindseth]{Hukkelas_2019}
Hukkel{\aa}s, H., Mester, R., and Lindseth, F.
\newblock Deepprivacy: A generative adversarial network for face anonymization.
\newblock In Bebis, G., Boyle, R., Parvin, B., Koracin, D., Ushizima, D., Chai,
  S., Sueda, S., Lin, X., Lu, A., Thalmann, D., Wang, C., and Xu, P. (eds.),
  \emph{Advances in Visual Computing}, pp.\  565--578, Cham, 2019. Springer
  International Publishing.
\newblock ISBN 978-3-030-33720-9.

\bibitem[{Isola} et~al.(2017){Isola}, {Zhu}, {Zhou}, and {Efros}]{Isola_2017}
{Isola}, P., {Zhu}, J., {Zhou}, T., and {Efros}, A.~A.
\newblock Image-to-image translation with conditional adversarial networks.
\newblock In \emph{2017 IEEE Conference on Computer Vision and Pattern
  Recognition (CVPR)}, pp.\  5967--5976, July 2017.
\newblock \doi{10.1109/CVPR.2017.632}.

\bibitem[Kingma \& Ba(2014)Kingma and Ba]{Kingma_2015}
Kingma, D.~P. and Ba, J.
\newblock Adam: A method for stochastic optimization, 2014.
\newblock URL \url{http://arxiv.org/abs/1412.6980}.
\newblock cite arxiv:1412.6980Comment: Published as a conference paper at the
  3rd International Conference for Learning Representations, San Diego, 2015.

\bibitem[Liu et~al.(2015)Liu, Luo, Wang, and Tang]{liu2015faceattributes}
Liu, Z., Luo, P., Wang, X., and Tang, X.
\newblock Deep learning face attributes in the wild.
\newblock In \emph{Proceedings of International Conference on Computer Vision
  (ICCV)}, December 2015.

\bibitem[Oh et~al.(2016{\natexlab{a}})Oh, Benenson, Fritz, and
  Schiele]{Oh_2016}
Oh, S.~J., Benenson, R., Fritz, M., and Schiele, B.
\newblock Faceless person recognition: Privacy implications in social media.
\newblock In Leibe, B., Matas, J., Sebe, N., and Welling, M. (eds.),
  \emph{Computer Vision -- ECCV 2016}, pp.\  19--35, Cham, 2016{\natexlab{a}}.
  Springer International Publishing.
\newblock ISBN 978-3-319-46487-9.

\bibitem[Oh et~al.(2016{\natexlab{b}})Oh, Benenson, Fritz, and
  Schiele]{oh2016faceless}
Oh, S.~J., Benenson, R., Fritz, M., and Schiele, B.
\newblock Faceless person recognition: Privacy implications in social media.
\newblock In \emph{European Conference on Computer Vision}, pp.\  19--35.
  Springer, 2016{\natexlab{b}}.

\bibitem[Oh et~al.(2017)Oh, Fritz, and Schiele]{oh2017adversarial}
Oh, S.~J., Fritz, M., and Schiele, B.
\newblock Adversarial image perturbation for privacy protection a game theory
  perspective.
\newblock In \emph{2017 IEEE International Conference on Computer Vision
  (ICCV)}, pp.\  1491--1500. IEEE, 2017.

\bibitem[Orekondy et~al.(2018)Orekondy, Fritz, and Schiele]{Orekondy_2018}
Orekondy, T., Fritz, M., and Schiele, B.
\newblock Connecting pixels to privacy and utility: Automatic redaction of
  private information in images.
\newblock In \emph{The IEEE Conference on Computer Vision and Pattern
  Recognition (CVPR)}, June 2018.

\bibitem[{Osia} et~al.(2020){Osia}, {Taheri}, {Shamsabadi}, {Katevas},
  {Haddadi}, and {Rabiee}]{Osia_2020}
{Osia}, S.~A., {Taheri}, A., {Shamsabadi}, A.~S., {Katevas}, K., {Haddadi}, H.,
  and {Rabiee}, H.~R.
\newblock Deep private-feature extraction.
\newblock \emph{IEEE Transactions on Knowledge and Data Engineering},
  32\penalty0 (1):\penalty0 54--66, Jan 2020.
\newblock ISSN 2326-3865.
\newblock \doi{10.1109/TKDE.2018.2878698}.

\bibitem[Raval et~al.(2017)Raval, Machanavajjhala, and
  Cox]{raval2017protecting}
Raval, N., Machanavajjhala, A., and Cox, L.~P.
\newblock Protecting visual secrets using adversarial nets.
\newblock In \emph{2017 IEEE Conference on Computer Vision and Pattern
  Recognition Workshops (CVPRW)}, pp.\  1329--1332. IEEE, 2017.

\bibitem[Ren et~al.(2018)Ren, Lee, and Ryoo]{Ren_2018}
Ren, Z., Lee, Y.~J., and Ryoo, M.~S.
\newblock Learning to anonymize faces for privacy preserving action detection.
\newblock In Ferrari, V., Hebert, M., Sminchisescu, C., and Weiss, Y. (eds.),
  \emph{Computer Vision -- ECCV 2018}, pp.\  639--655, Cham, 2018. Springer
  International Publishing.
\newblock ISBN 978-3-030-01246-5.

\bibitem[Ronneberger et~al.(2015)Ronneberger, P.Fischer, and
  Brox]{Ronneberger_2015}
Ronneberger, O., P.Fischer, and Brox, T.
\newblock U-net: Convolutional networks for biomedical image segmentation.
\newblock In \emph{Medical Image Computing and Computer-Assisted Intervention
  (MICCAI)}, volume 9351 of \emph{LNCS}, pp.\  234--241. Springer, 2015.
\newblock (available on arXiv:1505.04597 [cs.CV]).

\bibitem[Roy \& Boddeti(2019)Roy and Boddeti]{Roy_2019}
Roy, P.~C. and Boddeti, V.~N.
\newblock Mitigating information leakage in image representations: A maximum
  entropy approach.
\newblock In \emph{The IEEE Conference on Computer Vision and Pattern
  Recognition (CVPR)}, June 2019.

\bibitem[Salimans et~al.(2016)Salimans, Goodfellow, Zaremba, Cheung, Radford,
  Chen, and Chen]{Salimans_2016}
Salimans, T., Goodfellow, I., Zaremba, W., Cheung, V., Radford, A., Chen, X.,
  and Chen, X.
\newblock Improved techniques for training gans.
\newblock In Lee, D.~D., Sugiyama, M., Luxburg, U.~V., Guyon, I., and Garnett,
  R. (eds.), \emph{Advances in Neural Information Processing Systems 29}, pp.\
  2234--2242. Curran Associates, Inc., 2016.

\bibitem[Song et~al.(2017)Song, Lu, He, Sun, and Tan]{Song_2017}
Song, L., Lu, Z., He, R., Sun, Z., and Tan, T.
\newblock Geometry guided adversarial facial expression synthesis.
\newblock \emph{CoRR}, abs/1712.03474, 2017.
\newblock URL \url{http://arxiv.org/abs/1712.03474}.

\bibitem[Tang et~al.(2019)Tang, Xu, Liu, Wang, Sebe, and Yan]{Tang_2019}
Tang, H., Xu, D., Liu, G., Wang, W., Sebe, N., and Yan, Y.
\newblock Cycle in cycle generative adversarial networks for keypoint-guided
  image generation.
\newblock In \emph{Proceedings of the 27th ACM International Conference on
  Multimedia}, MM ’19, pp.\  2052–2060, New York, NY, USA, 2019.
  Association for Computing Machinery.
\newblock ISBN 9781450368896.
\newblock \doi{10.1145/3343031.3350980}.
\newblock URL \url{https://doi.org/10.1145/3343031.3350980}.

\bibitem[Tran et~al.(2018)Tran, Yin, and Liu]{tran2018representation}
Tran, L.~Q., Yin, X., and Liu, X.
\newblock Representation learning by rotating your faces.
\newblock \emph{IEEE transactions on pattern analysis and machine
  intelligence}, 2018.

\bibitem[Wang et~al.(2019)Wang, Wu, Wang, Wang, and Jin]{Wang_2019}
Wang, H., Wu, Z., Wang, Z., Wang, Z., and Jin, H.
\newblock Privacy-preserving deep visual recognition: An adversarial learning
  framework and {A} new dataset.
\newblock \emph{CoRR}, abs/1906.05675, 2019.
\newblock URL \url{http://arxiv.org/abs/1906.05675}.

\bibitem[Wu et~al.(2018)Wu, Wang, Wang, and Jin]{Wu_2018}
Wu, Z., Wang, Z., Wang, Z., and Jin, H.
\newblock Towards privacy-preserving visual recognition via adversarial
  training: A pilot study.
\newblock In Ferrari, V., Hebert, M., Sminchisescu, C., and Weiss, Y. (eds.),
  \emph{Computer Vision -- ECCV 2018}, pp.\  627--645, Cham, 2018. Springer
  International Publishing.
\newblock ISBN 978-3-030-01270-0.

\bibitem[Xiao et~al.(2020)Xiao, Tsai, Sohn, Chandraker, and Yang]{Xiao2020}
Xiao, T., Tsai, Y., Sohn, K., Chandraker, M., and Yang, M.
\newblock Adversarial learning of privacy-preserving and task-oriented
  representations.
\newblock In \emph{The Thirty-Fourth {AAAI} Conference on Artificial
  Intelligence, {AAAI} 2020, The Thirty-Second Innovative Applications of
  Artificial Intelligence Conference, {IAAI} 2020, The Tenth {AAAI} Symposium
  on Educational Advances in Artificial Intelligence, {EAAI} 2020, New York,
  NY, USA, February 7-12, 2020}, pp.\  12434--12441. {AAAI} Press, 2020.
\newblock URL \url{https://aaai.org/ojs/index.php/AAAI/article/view/6930}.

\bibitem[Xie et~al.(2017)Xie, Dai, Du, Hovy, and Neubig]{xie2017controllable}
Xie, Q., Dai, Z., Du, Y., Hovy, E., and Neubig, G.
\newblock Controllable invariance through adversarial feature learning.
\newblock In \emph{Advances in Neural Information Processing Systems}, pp.\
  585--596, 2017.

\bibitem[Zhang et~al.(2018)Zhang, Lemoine, and Mitchell]{zhang2018mitigating}
Zhang, B.~H., Lemoine, B., and Mitchell, M.
\newblock Mitigating unwanted biases with adversarial learning.
\newblock In \emph{Proceedings of the 2018 AAAI/ACM Conference on AI, Ethics,
  and Society}, pp.\  335--340. ACM, 2018.

\end{thebibliography}
\bibliographystyle{icml2021}

\newpage
\appendix
\section{Appendix}
\subsection{Additional details}
\label{additional_details}

\begin{figure}[]
  \centering
  \includegraphics[width=0.5\textwidth]{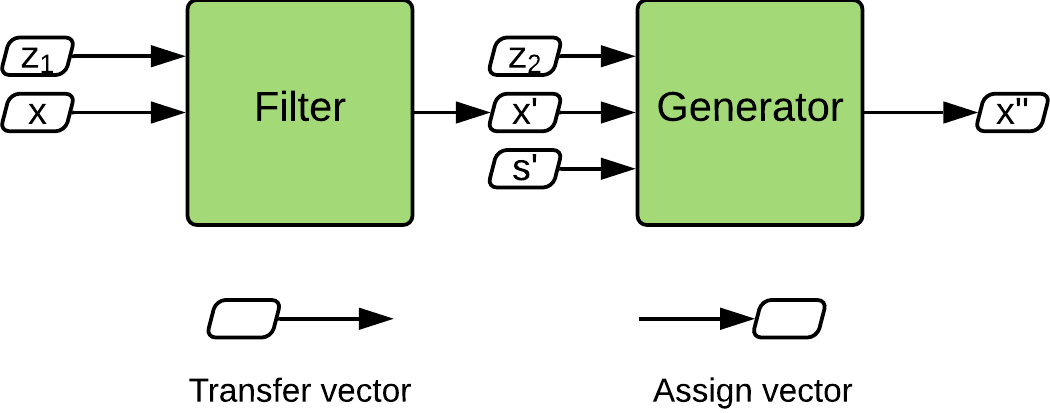} 
  \includegraphics[width=0.4\textwidth]{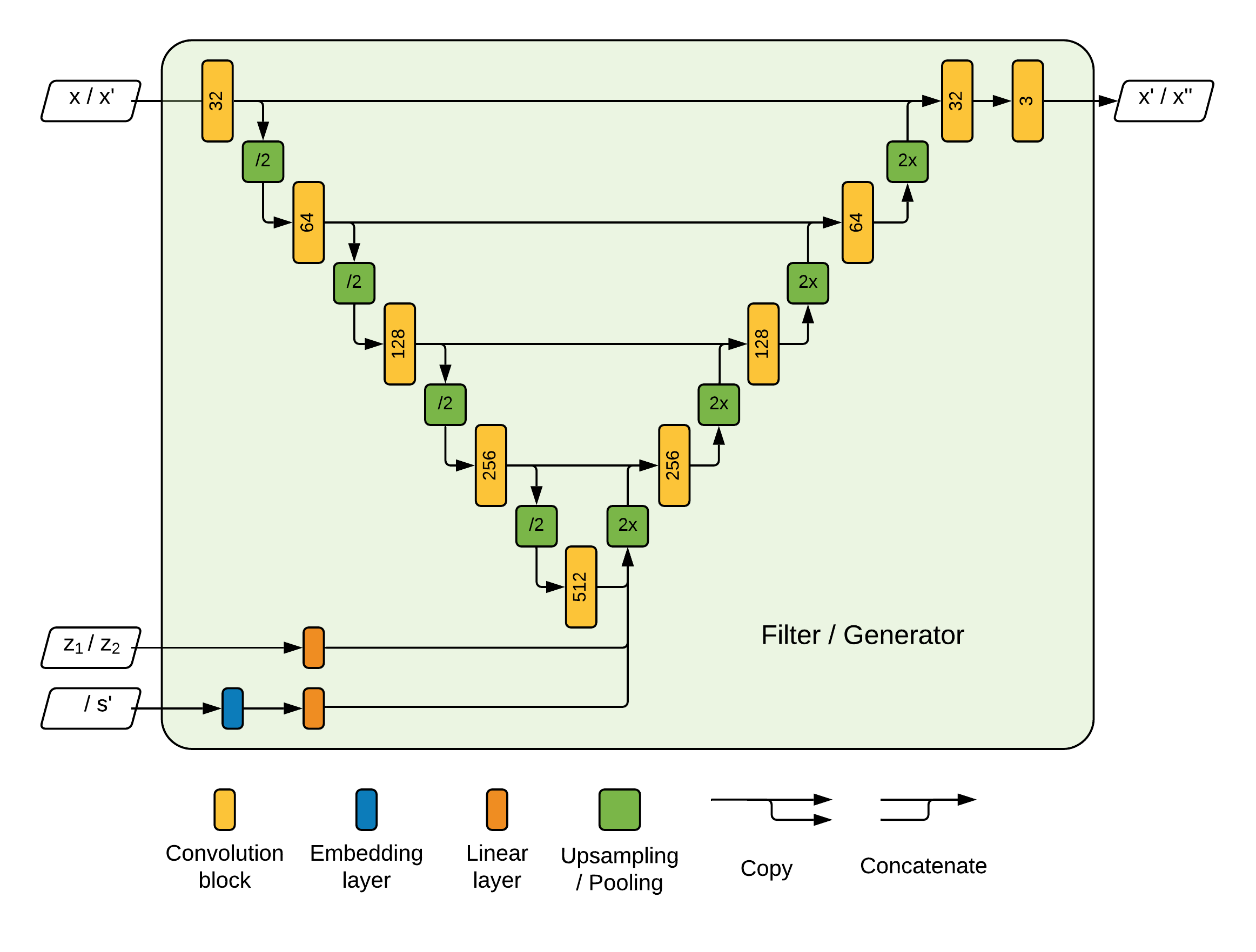}
  \caption{Overview of the training setup (left) and the network architecture used
  in the filter and the generator (right).}
  \label{overview}
\end{figure}

An overview of the setup can be seen in Figure~\ref{overview}. We use the
UNet~\cite{Ronneberger_2015}, illustrated on the right in Figure~\ref{overview}, architecture for
both the filter and the generator. The orange blocks are convolution blocks each
of which, except for the last block, consist of a convolution layer, a batch
normalization layer and a rectified linear activation unit, repeated twice in
that order. The number of output channels of the convolution layers in each
block has been noted in the figure. The last convolution block with a 3
channel output (the RGB image) consists of only a single convolutional layer
followed by a sigmoid activation. The green blocks denote either a max pooling
layer with a kernel size of two and a stride of two if marked with ``/2'' or a
nearest neighbor upsampling by a factor of two if marked with ``2x''. The blue
block denotes an embedding layer, which takes as input the categorical value of
the sensitive attribute and outputs a dense embedding of 128 dimensions. It is
then followed by a linear projection and a reshaping to match the spatial
dimensions of the output of the convolution block to which it is concatenated,
but with a single channel. The same type of linear projection is applied on the
1024 dimensional noise vector input, but this projection and reshaping matches
both the spatial and channel dimensions of the output of the convolutional block
to which it is concatenated. Concatenation is in both cases done along the
channel dimension.

\subsection{Additional results}
\label{additional_results}
The results in Table~\ref{tab:results_2} show that using the filter together
with the generator ($g \circ f$) give a stronger privatization guarantee than using only the
generator (without the filter) or only the baseline (the filter without the generator).

\begin{table}[h]
  \caption{The results of evaluating the adversarially trained classifiers
    $adv(s|\cdot)$ on the held out test data
  censored with the baseline, only the generator, and our method
  for varying distortion $\epsilon$ for the \textit{smiling} attribute. Closer to 50 means more
privacy. Only using the generator does not privide any privacy, but using it
together with the filter provides the best privacy. Lower is better.}
  \label{tab:results_2}
\begin{center}
\begin{tabular}{llll}
  \textbf{Dist.} &\multicolumn{3}{c}{\textbf{Adversarial smiling}}  \\
  $\epsilon$ & baseline ($f$) & generator ($g$) & ours  ($g\circ f$)\\
\hline \\
  0.001 & 89.3 & 90.7 & 88.7 \\
  0.005 & 89.9 & 91.2 & 85.9 \\
  0.01  & 89.8 & 91.5 & 83.2 \\
  0.05  & 69.2 & 78.7 & 54.1 \\
\end{tabular}
\end{center}
\end{table}

\begin{figure*}[]
  \centering
  \includegraphics[width=1.0\textwidth]{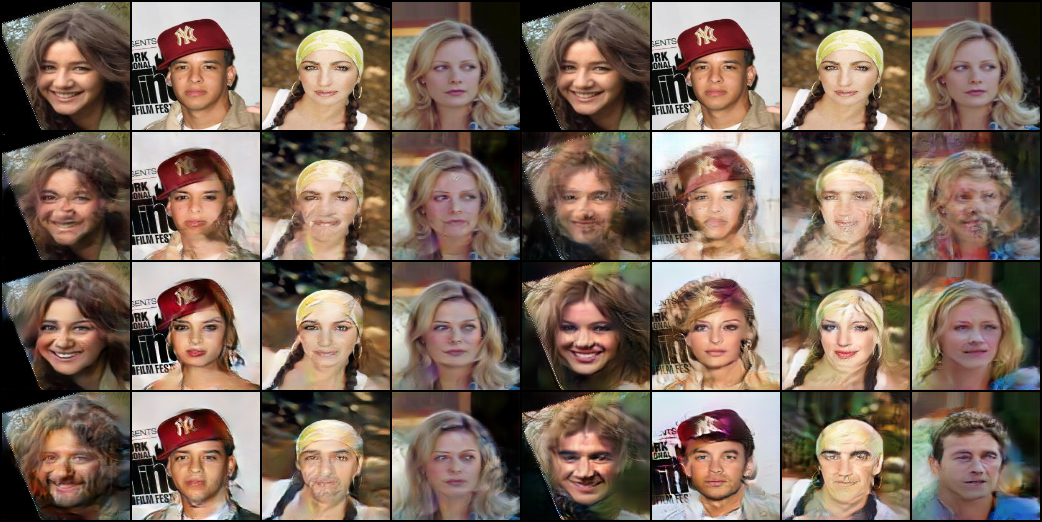}
  \caption{Qualitative results for the sensitive attribute gender. In the first four columns: $\epsilon=0.001$,
    and  in the last four columns: $\epsilon=0.01$. From top to bottom row: input image ($x$), censored image ($x'$), censored image
    with synthetic female gender ($x'', s'=0$), censored image with synthetic male gender 
    ($x'', s'=1$). The model is able to generate a synthetic gender
	while maintaining much of the structure in the image. These images were
generated from a model trained using 128x128 pixels.} 
  \label{fig:qualitative_2}
\end{figure*}

In Figure~\ref{fig:qualitative_2} we show additional qualitative results when
the attribtue \textit{gender} is considered sensitive.

Table~\ref{tab:results_5} shows the correlations between classifier predictions
on a pair of attributes when one attribute has been synthetically replaced.

For example, in the CelebA dataset lipstick is highly correlated with female
gender.  This means that if we want to hide all information about whether or not
the person is wearing lipstick we also need to hide its gender (and other
correlating attributes). This problem can be seen in Table~\ref{tab:results_5}
where changing whether or not a person is wearing lipstick correlates with
changes of gender.

The question is: if we censor an attribute in an image, how does that correlate
with changes of other attributes in the image? In the lipstick column of
Table~\ref{tab:results_5} we have censored the attribute lipstick. We then make
predictions on whether or not the person in the censored image is wearing
lipstick, and compute the correlation between these predictions and predictions
for the attributes for each row. We can see that changes in lipstick correlate
negatively with changes in gender and positively with makeup.  This highlights
the problem of disentangling these underlying factors of variation. 

\begin{table}[]
  \caption{The value of each
    cell denotes the Pearson's correlation coefficient between predictions from a classifier trained
    to predict the row attribute and a classifier trained to predict the
    column attribute, given that the column attribute has been censored.} 
  \label{tab:results_5}
\begin{center}
\setlength{\tabcolsep}{5pt} 
\begin{tabular}{rrrrr}
  & Smiling & Gender & Lipstick & Young \\
\hline \\
             Smiling &                 1.00 &                -0.04 &                 0.08 &                -0.06 \\
                Gender &                -0.07 &                 1.00 &                -0.44 &                -0.21 \\
    Lipstick &                 0.14 &                -0.30 &                 1.00 &                 0.26 \\
\vspace{0.5em}               Young &                 0.05 &                -0.11 &                 0.23 &                 1.00 \\
     High Cheekbones &                 0.14 &                -0.07 &                 0.15 &                -0.01 \\
 Mouth Open &                 0.04 &                 0.00 &                 0.03 &                -0.02 \\
        Heavy Makeup &                 0.12 &                -0.24 &                 0.47 &                 0.22 \\
\end{tabular}
\end{center}
\end{table}

Table~\ref{tab:correlations} shows the Pearson correlations between the smiling attribute
and 37 other attributes in the CelebA dataset.

\begin{table}[t]
\caption{Pearson correlation coefficient between the smiling attribute and 37 other attributes
  in the CelebA training dataset, ordered from high to low absolute correlation.} 
\label{tab:correlations}
\begin{center}
\begin{tabular}{lr}
  Attribute & Correlation \\
\hline \\
High cheekbones &        0.68 \\
Mouth slightly open &    0.53 \\
Rosy cheeks &            0.22 \\
Oval face &              0.21 \\
Wearing lipstick &       0.18 \\
Heavy makeup &           0.18 \\
Wearing earrings &       0.17 \\
Attractive &             0.15 \\
Gender &                -0.14 \\
Bags under eyes &        0.11 \\
No beard &               0.11 \\
Big nose &               0.10 \\
Double chin &            0.10 \\
Arched eyebrows &        0.09 \\
Wearing necklace &       0.09 \\
Blond hair &             0.09 \\
Narrow eyes &            0.08 \\
Sideburns &             -0.08 \\
Wavy hair &              0.08 \\
Mustache &              -0.07 \\
Pale skin &             -0.07 \\
Five-o'clock shadow &   -0.07 \\
Goatee &                -0.07 \\
Blurry &                -0.06 \\
Wearing hat &           -0.06 \\
Bangs &                  0.05 \\
Chubby &                 0.04 \\
Eyeglasses &            -0.04 \\
Pointy nose &            0.04 \\
Brown hair &             0.02 \\
Receding hairline &      0.02 \\
Bald &                   0.01 \\
Big lips &               0.01 \\
Gray hair &              0.01 \\
Straight hair &          0.01 \\
Black hair &            -0.00 \\
Bushy eyebrows &        -0.00 \\
Wearing necktie &       -0.00 \\
\end{tabular}
\end{center}
\end{table}

\subsection{Training Setup}
\label{sec:training_setup}
Algorithm \ref{algorithm} demonstrates the training setup for our method. Note
that when $\ell_f$ is negative entropy it does not depend on the $1-s_i$
argument, but is simply computed from the output of the discriminator $h_f$.
However, when $\ell_f$ is cross entropy we optimize the filter such that the discriminator
$h_f$ is fooled into predicting the complement class $1-s$. This means that we need to
assume $s$ to be a binary attribute when using this loss. When using the
negative entropy we do not need to make this assumption, which is a
strength of the proposed method.
\begin{algorithm}[h]
  \caption{}
  \label{algorithm}
  \begin{algorithmic} 
    \STATE{\bfseries input:} $\mathcal{D}, lr, \lambda, \epsilon, \beta_1, \beta_2$
    \STATE $\epsilon_1, \epsilon_2 \gets \epsilon$
    \REPEAT 
    \STATE Draw m samples uniformly at random from the dataset
    \STATE{$(x_1, s_1),\dots,(x_m, s_m) \sim \mathcal{D}$}
    \STATE{Draw m samples from the noise distribution}
    \STATE{$(z^{(1)}_1, z^{(2)}_1), \dots, (z^{(1)}_m, z^{(2)}_m) \sim p(z^{(1)}, z^{(2)})$}
    \STATE{Draw m samples from the synthetic distribution}
    \STATE{$s'_1, \dots, s'_m \sim p(s')$}
    \STATE{Compute censored and synthetic data}
    \STATE{$x'_1, \dots, x'_m = f_{\theta_f}(x_1, z^{(1)}_1), \dots, f_{\theta_f}(x_m,z^{(1)}_m)$}
    \STATE{$x''_1, \dots, x''_m = g_{\theta_g}(x'_1, s'_1, z^{(2)}_1), \dots, g_{\theta_g}(x'_m, s'_m, z^{(2)}_m)$}
    \STATE{Compute filter and generator losses}
    \STATE{$\begin{aligned}\Theta_f(\theta_f) &=
    \frac{1}{m}\sum_{i=1}^{m}\ell_f(h_f(x'_i;\phi_f), 1-s_i) \\ &+
    \lambda\max(\frac{1}{m}\sum_{i=1}^{m}d(x'_i, x_i)-\epsilon, 0)^2\end{aligned}$}
    \STATE{$\begin{aligned}\Theta_g(\theta_g) &=
      \frac{1}{m}\sum_{i=1}^{m}\ell_g(h_g(x''_i;\phi_g), s'_i) \\
        &+ \lambda\max(\frac{1}{m}\sum_{i=1}^{m}d(x''_i, x_i)-\epsilon,
    0)^2\end{aligned}$}
    \STATE{Update filter and generator parameters}
    \STATE{$\theta_f \gets \text{Adam}(\Theta_f(\theta_f); lr, \beta_1, \beta_2$)}
    \STATE{$\theta_g \gets \text{Adam}(\Theta_g(\theta_g); lr, \beta_1, \beta_2$)}
    \STATE{Compute discriminator losses}
    \STATE{$\Phi_f(\phi_f) = \frac{1}{m}\sum_{i=1}^{m}\ell_{h_f}(h_f(x'_i;\phi_f), s_i)$}
    \STATE{$\begin{aligned}\Phi_g(\phi_g) &= \frac{1}{m}\sum_{i=1}^{m}\ell_g(h_g(x''_i;\phi_g),
      fake) \\ &+ \frac{1}{m}\sum_{i=1}^{m}\ell_g(h_g(x_i;\phi_g), s_i)\end{aligned}$}
    \STATE{Update discriminator parameters}
    \STATE{$\phi_f \gets \text{Adam}(\Phi_f(\phi_f); lr, \beta_1, \beta_2$)}
    \STATE{$\phi_g \gets \text{Adam}(\Phi_g(\phi_g); lr, \beta_1, \beta_2$)}
    \UNTIL{stopping criterion}
  \STATE{\bfseries return} $\theta_f, \theta_g, \phi_f, \phi_g$
  \end{algorithmic}
\end{algorithm}


\end{document}